\documentclass[runningheads]{llncs}
\usepackage{graphicx}

\usepackage{tikz}
\usepackage{comment}
\usepackage{amsmath,amssymb} 
\usepackage{color}

\usepackage[accsupp]{axessibility}  

\usepackage{graphicx}

\graphicspath{{Figures/}}

\begin{document}
\pagestyle{headings}
\mainmatter
\def\ECCVSubNumber{4368}  

\title{StyleFlow For Content-Fixed Image to Image Translation} 

%
\author{Weichen Fan\thanks{Equal contribution.} \and
Jinghuan Chen$^{*}$ \and Jiabin Ma \and Jun Hou \and Shuai Yi}

\authorrunning{Fan et al.}
%
\institute{Sensetime Research}
\maketitle

\begin{abstract}
Image-to-image (I2I) translation is a challenging topic in computer vision. We divide this problem into three tasks: strongly constrained translation, normally constrained translation, and weakly constrained translation. The constraint here indicates the extent to which the content or semantic information in the original image is preserved. Although previous approaches have achieved good performance in weakly constrained tasks, they failed to fully preserve the content in both strongly and normally constrained tasks, including photo-realism synthesis, style transfer, and colorization, .etc. To achieve content-preserving transfer in strongly constrained and normally constrained tasks, we propose StyleFlow, a new I2I translation model that consists of normalizing flows and a novel Style-Aware Normalization (SAN) module. With the invertible network structure, StyleFlow first projects input images into deep feature space in the forward pass, while the backward pass utilizes the SAN module to perform content-fixed feature transformation and then projects back to image space. Our model supports both image-guided translation and multi-modal synthesis. We evaluate our model in several I2I translation benchmarks, and the results show that the proposed model has advantages over previous methods in both strongly constrained and normally constrained tasks.
\end{abstract}
\section{Introduction}
\label{sec:intro}
\begin{figure}
    \centering
    \includegraphics[width=1.0\linewidth]{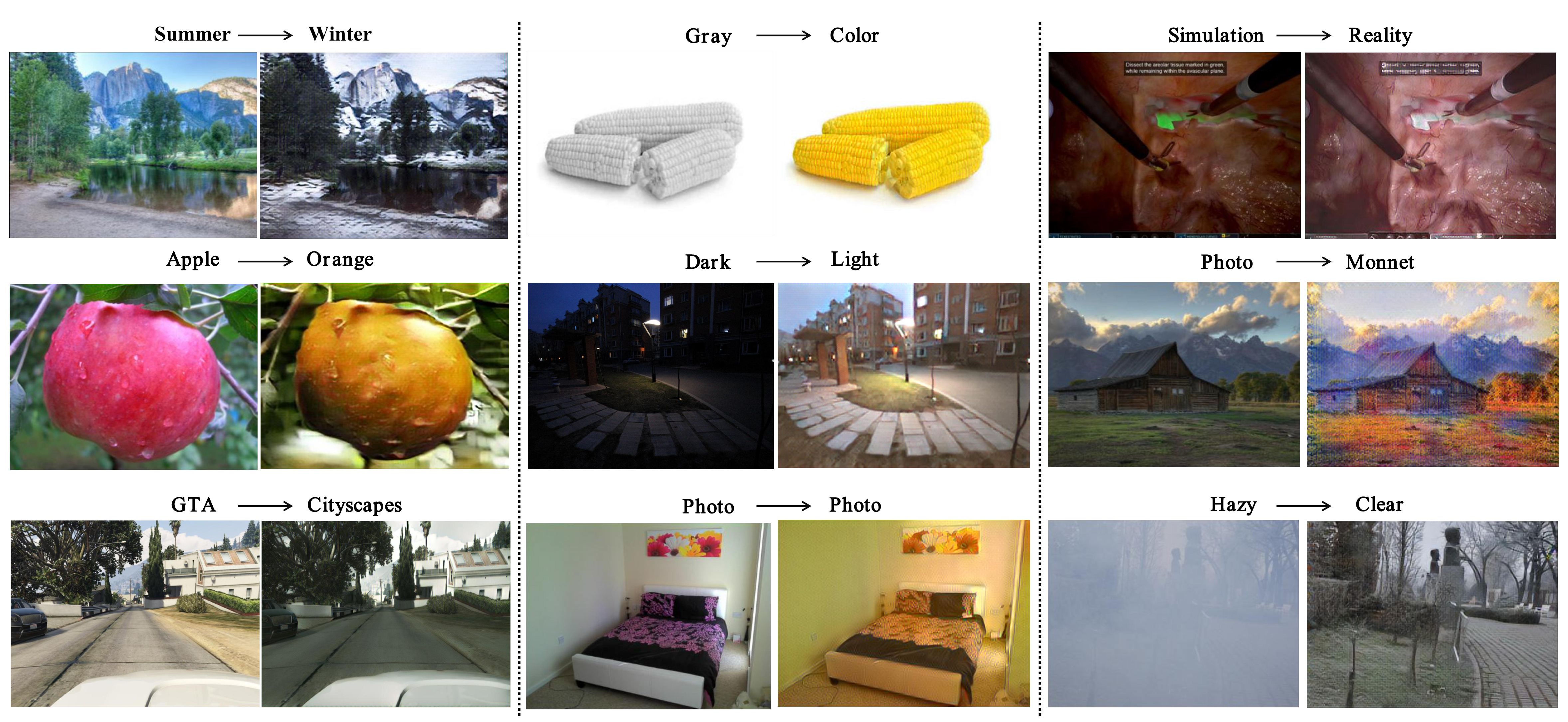}
    \caption{Given two images in different visual domains, our model learns to translate from one to the other in various tasks.(1. Summer to Winter\cite{zhu2017unpaired}; 2. Gray to Color\cite{anwar2020image}; 3. Sim to Real\cite{zia2021surgical}; 4. Apple to Orange\cite{zhu2017unpaired}; 5. Dark to Light\cite{wei2018deep}; 6. Photo to Monnet\cite{wikiart}; 7. GTA\cite{richter2016playing} to Cityscapes\cite{cordts2016cityscapes}; 8. Photo to Photo\cite{luan2017deep}; 9. Hazy to Clear\cite{Dense-Haze_2019,NTIRE_Dehazing_2019})}
    \label{fig:various_results}
\end{figure}

Image-to-image (I2I) translation\cite{isola2017image} is a long-standing topic in computer vision, which is required to learn a mapping between two different visual domains while preserving the semantic information (content) of the source domain and obtaining the domain properties (style) of the target domain. Many applications, such as style transfer\cite{gatys2015texture,Gatys_2016_CVPR,huang2017arbitrary,li2017universal}, super-resolution\cite{Ledig_2017_CVPR,Lai_2017_CVPR}, image enhancement\cite{lore2017llnet} and photo-realism synthesis\cite{Chen_2017_ICCV,Wang_2018_CVPR,richter2021enhancing}, can be formulated as I2I translation problems. According to the requirement of content preservation during translation, these problems can be further divided into three subsets: strongly constrained translation, normally constrained translation, and weakly constrained translation.

Recent methods\cite{zhu2017unpaired,huang2018multimodal,lee2018diverse} based on cycle-consistency constraint have shown great progress in I2I translation, by forcing the translated images to be mapped back to the original images during training. Although these methods have achieved impressive visual results in most applications, the cycle-consistency constraint failed to reproduce rich and complex semantic information, thus resulting in different levels of content distortion in translated images, especially in normally and strongly constrained translation settings.

In this work, we propose \textit{StyleFlow} to address the problem of content distortion in I2I translation. Unlike previous methods which follow \textit{encoder-decoder} scheme, StyleFlow utilizes the design of normalizing flow models\cite{dinh2014nice}, which consists of a series of fully invertible basic blocks, to achieve lossless image reconstruction. In addition, a novel \textit{Style-Aware Normalization (SAN)} module is introduced to perform context-fixed feature transformation. Given source image features and target image features, SAN adjusts mean and variance of source features with learnable \textit{content-guided affine parameters} to match the style of target input while preserve the content of source input. Considering that our model only consists of fully reversible transformation during feature extraction and reconstruction, pixel-level reconstruction loss is not required. Following most I2I translation tasks especially in style transfer\cite{gatys2015texture}, content loss and style loss calculated based on a pre-trained VGG encoder\cite{Simonyan2015very} are adopted. We further extend the idea of style loss and introduce its simple extension named \textit{aligned-style loss}, which takes the trade-off between content preservation and stylization into consideration, to further improve translation results especially when unpaired training images are provided.

We apply the proposed framework to a wide range of applications, plausible results (see Figure \ref{fig:various_results}) indicate the significance and effectiveness of the designed modules in our method. We summarize the contributions of this work as below: 1) We divide image-to-image translation tasks into three subsets: strongly, normally and weakly constrained translation, according to the requirement of content preservation. 2) We propose StyleFlow, a novel model based on invertible network structure for content-fixed image-to-image translation, which is capable for unpair, multi-modal and multi-domain translation settings. 3) We design a novel Style-Aware Normalization module for efficient content-fixed feature transformation. 4) We demonstrate that StyleFlow outperforms previous methods with high content preservation and admirable stylization in extensive experiments.
\begin{table}[!h]
  \centering
  \caption{Feature comparison of I2I models.}
  \label{tab:comp}
  \begin{tabular}{|c|c|c|c|c|c|}
    \hline
    Method &  Unpaired & Multimodal & Multi-domain & Invertible & Parameters\\
    \hline
    \textbf{StyleFlow} &  \checkmark & \checkmark & \checkmark & \checkmark & 16.78 M\\
    \hline
    DRIT++\cite{lee2020drit++} & \checkmark &\checkmark & \checkmark& & 18.58 M\\
    \hline
    MUNIT\cite{huang2018multimodal} & \checkmark &\checkmark & & & 30.05 M\\
    \hline
    UNIT\cite{liu2017unsupervised} & \checkmark & & & & 22.28 M\\
    \hline
    CycleGAN\cite{zhu2017unpaired} & \checkmark & & & & 11.38 M\\
    \hline
    CUT\cite{park2020contrastive} & \checkmark & & & & 11.38 M\\
    \hline
  \end{tabular}
\end{table}
\section{Related Work}
\label{sec:related}
\subsection{Image-to-Image Translation}
\noindent\textbf{Generic Image-to-Image Translation}: The previous generic I2I models can be divided into two categories: VAE-based\cite{ma2018exemplar,zhu2017multimodal,lee2018diverse,liu2017unsupervised,huang2018multimodal,wu2019transgaga} and GAN-based\cite{park2020contrastive,zhu2017unpaired,kim2017learning,yi2017dualgan,chen2020reusing}. However, both of them surfer from the problem of content distortion, even though a lot of regularization methods have been proposed to reduce the impact, including cyclic consistency, self-reconstruction, etc. For weakly constrained translations, where the content can be heavily modified, these methods are appropriate, while for strongly constrained and normally constrained translations, they do not perform well. Our proposed model can be applied to both strongly constrained and normally constrained translations without the problem of content distortion.\\[1ex]
\noindent\textbf{Strongly Constrained Translation}: Strongly constrained I2I translation means the content of the source image should be preserved to a great extent. Photo-realism\cite{richter2021enhancing,hoffman2018cycada,cherian2019sem}, Colorization\cite{kumar2021colorization,deshpande2017learning,guadarrama2017pixcolor,cao2017unsupervised}, and Image Enhancement\cite{lore2017llnet,wei2018deep,jiang2021enlightengan} belong to the strongly constrained setting. These problems require the translated images to retain the exact rich and complex content information in source images. To achieve high content preservation, previous methods require paired training images or auxiliary inputs such as semantic segmentation masks.\\[1ex]
\noindent\textbf{Normally Constrained Translation}: Normally constrained I2I translation contains style transfer\cite{gatys2015texture,huang2017arbitrary,sheng2018avatar,gu2018arbitrary,wang2020diversified,chen2016fast,li2017universal,an2020ultrafast}, season and weather transfer\cite{li2021weather}, raindrop removal\cite{qian2018attentive,shao2021selective}, haze removal\cite{engin2018cycle,anvari2020dehaze,fahim2021single}, and image denoising\cite{guo2019toward,shan2019residual,lehtinen2018noise2noise}. In this setting, the source and target domains usually show different visual effects, such as weather conditions and artistic styles, but share similar structural information, the primary objective is to transfer the overall visual effects of source domains to match those in target domains. Previous work have shown plausible overall visual results in these tasks, while certain level of content distortion can be found when we zoom in to the details of translated images.\\[1ex]
\noindent\textbf{Weakly Constrained Translation}: Weakly constrained I2I translation refers to problems where the source and target images may lie in completely different domains or modals, the translation is to be performed on a high semantic level, which means the content information of source images can be modified a lot. Label to image\cite{lin2018conditional,isola2017image} and object to object translation\cite{huang2018multimodal,lee2018diverse,liu2017unsupervised,zhu2017unpaired,park2020contrastive} belong to this type of problem.

\subsection{Normalizing Flow}
Normalizing flow is a type of  generative model that uses a sequence of invertible mappings to transform from distribution to distribution, and it is accurate and efficient in both density estimation and sampling\cite{kobyzev2020normalizing}. Dinh et al.\cite{dinh2014nice} first propose a flow-based generative model, NICE. After that, GLOW\cite{kingma2018glow}, RealNVP\cite{dinh2016density}, FLOW++\cite{ho2019flow++} are proposed to improve the sample efficiency and density estinmation performance. More recently, BeautyGLOW\cite{chen2019beautyglow} is proposed for makeup transfer. Besides, ArtFlow\cite{an2021artflow} proves that the normalizing flow is unbiased in neural style transfer compared with the previous work. Our model is based on GLOW\cite{kingma2018glow} to achieve invertible transformation. 
\section{Methodology}
\begin{figure*}
  \centering
  \includegraphics[width=1.0\linewidth]{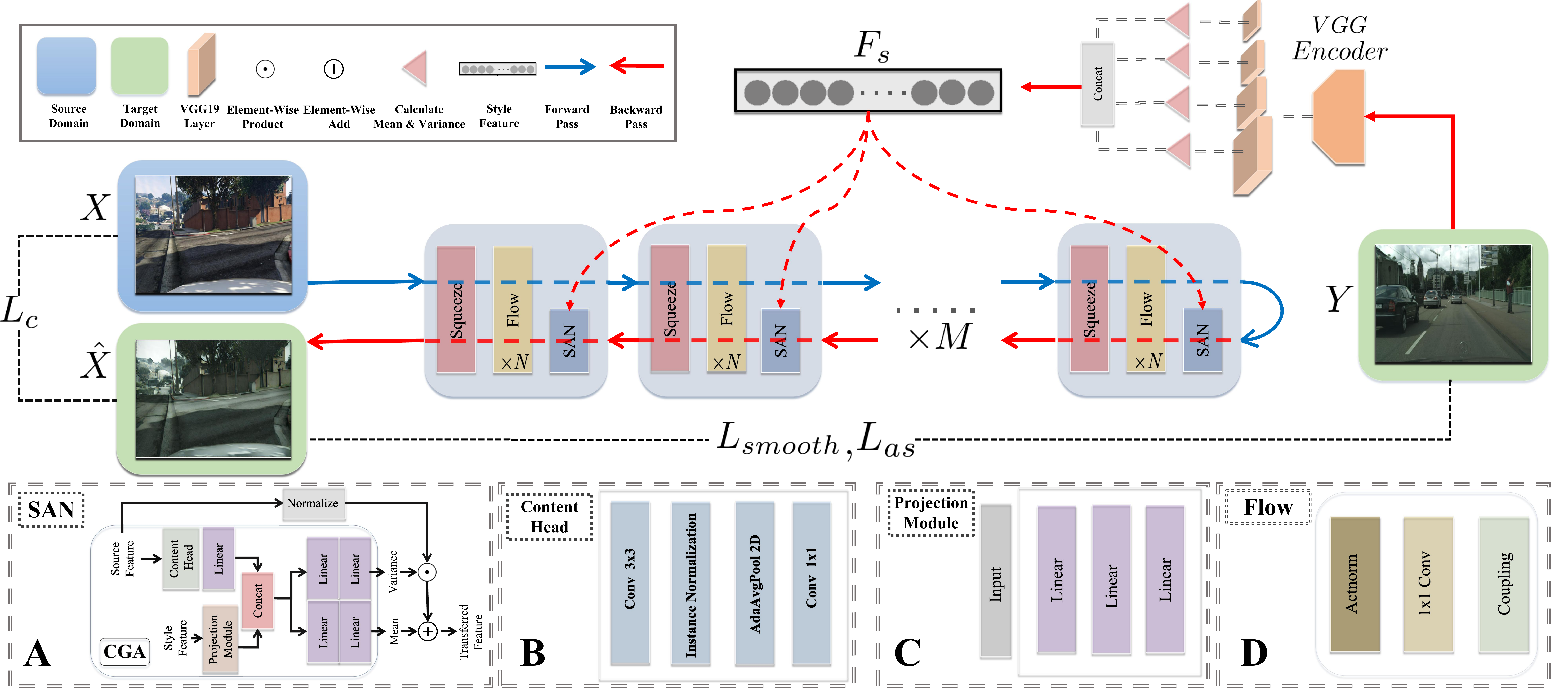}
  \caption{The framework of our proposed StyleFlow. Our model works in an invertible manner. The \textcolor{blue}{blue} arrows indicate the forward pass to extract features, while the \textcolor{red}{red} arrows represent the backward pass to reconstruct images. Our model consists of a series of reversible blocks, where each block has three components: the squeeze module, the Flow module, and the SAN module. A pre-trained VGG encoder is used for domain features extraction and loss computation.}
  \label{fig:overview}
\end{figure*}

As shown in the Figure \ref{fig:overview}, we present StyleFlow, which consists of invertible normalizing flows and a novel Style-Aware Normalization (SAN) module to achieve content-fixed image-to-image translation. Following previous works, a pretrained and fixed VGG-19 encoder $E_{VGG}$ is utilized in our method for feature extraction and loss computation. As our model is invertible, we denote the forward pass as $E$, and the backward pass as $E^{-1}$. Taking a source domain $X$ and a target domain $Y$ as an example. The forward pass of StyleFlow maps images from source domain into deep features, i.e. $E: X\rightarrow {F_X}$, where $F_X$ is source feature space. The pretrained VGG encoder would map the target domain into a shared style space, i.e. $E_{VGG}: Y\rightarrow F_S$. Taking $F_X$ and $F_S$ as inputs, SAN module performs content-fixed feature transformation to obtain features in the stylized space $\hat{F}_X$. And the backward pass of StyleFlow generates translated images by mapping the stylized features back to the image space, i.e. $E^{-1}: \hat{F}_X \rightarrow \hat{X}$, where $\hat{X}$ is assumed to share the style properties of $Y$ while retain the content information of $X$.

In this section, we will first give a analysis on invertible mapping in section \ref{sec:Invertible}, then discuss the details of our proposed method in \ref{sec:StyleFlow}, and lastly, the loss function design in \ref{sec:Loss}. The summary of components used in StyleFlow is shown in Table \ref{tab:cap}.


\begin{table}[h]
  \centering
  \caption{Summary of components used in StyleFlow.}
  \label{tab:cap}
  \begin{tabular}{|c|c|c|c|c|}
    \hline
    Module &  Unpaired & Multimodal & Multi-domain & Invertible\\
    \hline
    StyleFlow &  \checkmark & \checkmark & \checkmark & \checkmark\\
    \hline
    w/o SAN &  \checkmark &  &  & \checkmark\\
    \hline
    w/o Flow &  \checkmark &  &  \checkmark& \\
    \hline
    w/o Aligned-Style Loss & & \checkmark &  \checkmark& \checkmark\\
    \hline
  \end{tabular}
\end{table}

\subsection{Invertible Analysis}
\label{sec:Invertible}
\begin{figure}[h]
  \centering
  \includegraphics[width=0.7\linewidth,height=0.4\linewidth]{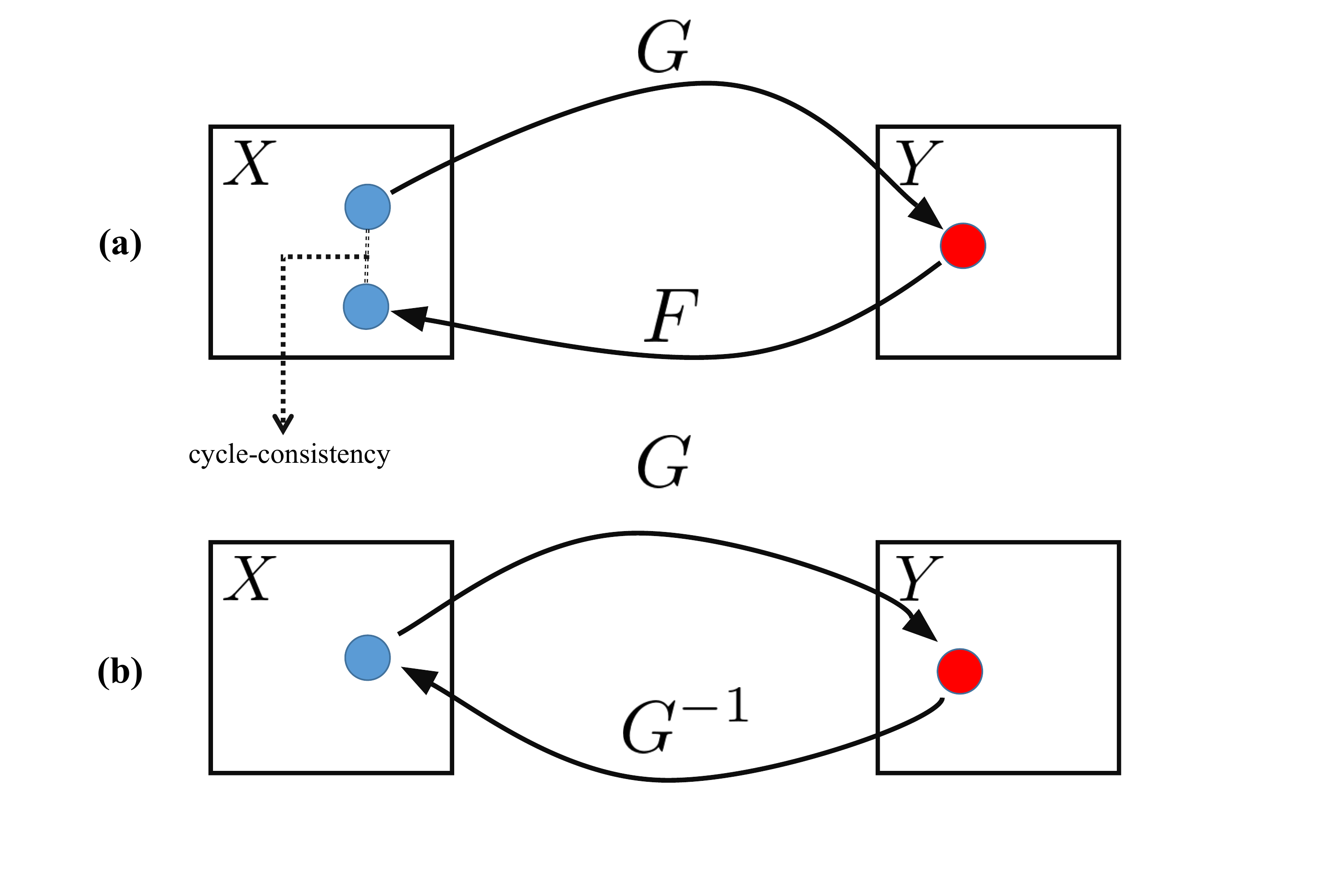}
  \caption{(a). Previous work try to make mapping $G$ and $F$ inverse to each other by using cycle-consistency loss. (b). Our model use a revertible mapping $G$ to map from $X$ to $Y$, and its inverted function $G^{-1}$ would map from $Y$ back to $X$ exactly.}
  \label{fig:cyc}
\end{figure}

As shown in Figure \ref{fig:cyc}, given source domain $X$ and target domain $Y$, previous works (CycleGAN\cite{zhu2017unpaired}, MUNIT\cite{huang2018multimodal}, DRIT\cite{lee2018diverse}, .etc.) use cycle consistency to regularize the forward-pass generator $G$ and backward-pass generator $F$ to become invertible in ideal situation. However, these models could not achieve fully-revertible mapping. To achieve bijective mapping, Behrmann\cite{behrmann2019invertible} et al. has proposed an invertible resnet under Lipschitz constrain. Other works\cite{dinh2014nice,dinh2016density,kingma2018glow} create invertible architecture based on carefully-designed coupling or auto-regressive structures. 

Based on nomalizing flow, in order to achieve a bijective mapping $f: X \leftrightarrow Y$, with a probability distribution $p$, we have the forward and backward transformation based on change of variable theory:
\begin{align}
p_X(x) &= p_Y(f(x))|det\frac{\partial f(x)}{\partial x}|\\
  p_Y(x) &= p_X(f(x))|det\frac{\partial f(x)}{\partial x}|^{-1}
\end{align}
where $x\in X$, $y\in Y$, and $\frac{\partial f(x)}{\partial x}$ is the jacobian of $f$ at $x$. To achieve invertible transformation, the determinant of the jacobian matrix should not be zero. Given a set of invertible transformation $g_1,g_2,...,g_n$, the $det|\frac{\partial g_i}{\partial_x}|$ ($\forall i \in [1,n]$) is non-zero. Therefore, $\prod_{i=1}^n det|\frac{\partial g_i}{\partial_x}|$ is non-zero, which means the composition of $g_1$ to $g_n$ will form a more complex invertible transformation. Further explanations are out of the scope of this work, we refer readers to previous works\cite{dinh2014nice,dinh2016density,kingma2018glow,kobyzev2020normalizing} for more details.
\subsection{StyleFlow}
\label{sec:StyleFlow}
Based on the effective invertible model GLOW\cite{kingma2018glow}, we proposed a novel model named StyleFlow for image to image translation. As shown in the Fig \ref{fig:overview}, our proposed model consists $M$ blocks, each block is made up of a squeeze operation, $N$ Flow Modules, and a novel Style-Aware Normalization (SAN) module. According to Sec \ref{sec:Invertible}, if each block is invertible, we can cascade multiple blocks for more complex reversible transformation.\\[1ex]
\noindent\textbf{A. Squeeze Operation} serves as the inter-connection between blocks for feature reorganization. It reduces spatial size of feature map by first splitting input feature into smaller patches along spatial dimension and then concatenating patches along channel dimension.\\[1ex]
\noindent\textbf{B. Flow Module} consists of three reversible transformations: Actnorm layer, 1x1 Convolution layer and Coupling layer. 

Actnorm refers to the activation normalization layer introduced in GLOW\cite{kingma2018glow}, which performs feature normalization along channel dimension. The forward and backward function can be written as:
\begin{align}
      y_{i,j} &= s\odot x_{i,j} + b\\
      x_{i,j} &= (y_{i,j} - b) / s
\end{align}
where $i,j$ represents the spatial position of input tensor, $s$ and $b$ refer to the learnable affine parameters scale and bias.

Coupling layer is first proposed in NICE\cite{dinh2014nice} to achieve efficient invertible transformation. Given an input feature $x\in \mathbb{R}^D$, output feature $y\in \mathbb{R}^D$ can be obtained with:
\begin{align}
      y_{1:d} &= x_{1:d}\\
      y_{d+1:D} &= g(x_{d+1:D}, m(x_{1:d}))
\end{align}
where $d$ splits $x$ into two disjoint partitions along channel dimension, and $g$ is called \textit{coupling function}. 
In our model, we define $g$ as a simple subtractive function:
\begin{align}
      y_{d+1:D} &= x_{d+1:D} - m(x_{1:d})
\end{align}
where $m$ could be any neural network with input channel $d$ and output channel $D-d$, as $m$ itself is not required to be invertible. With this design, the backward pass of the coupling layer can be easily derived as:
\begin{align}
    x_{1:d} &= y_{1:d}\\
    x_{d+1:D} &= y_{d+1:D} + m(x_{1:d})
\end{align}
In despite of the simplicity of coupling layer, extensive experiments show that this subtractive coupling function is sufficient to produce favourable performance in our case.

Since the coupling layer splits input feature into two partitions, only one part of the feature is modified during transformation. To avoid this, we use the invertible 1x1 convolution layer in GLOW\cite{kingma2018glow} for channel permutation.\\[1ex]
\noindent \textbf{C. Style-Aware Normalization Module}

We proposed a novel Style-Aware Normalization (SAN) module to achieve content-fixed feature transformation. As the results shown in \cite{huang2017arbitrary}, in I2I translation, content and style are both spatial statistical information of images. To be more specific, content is the information that we would like to preserve during the transformation (e.g. shape, semantic information), while style is what we need to change to make the source image 'similar' to the target image (e.g. color, illumination, clarity). Content-fixed transfer means the content information before and after transformation should be retained.\\[1ex]
\noindent\textbf{Content \& Style}:
Following conventional definitions, given the extracted features in latent space of an image, content is defined as the channel-wise normalized term, while style can be expressed with its mean ($\mu$) and variance ($\sigma$). In this work, we use normalizing flows to extract source images features and pre-trained VGG encoder to extract target image features. Thus, given source image $x$ and target image $y$, content features $f_c$ and style features $f_s$ can be expressed as below:
\begin{align}
    f_c &= \frac{f_x - \mu(f_x)}{\sigma(f_x)} = \frac{E(x) - \mu(E(x))}{\sigma(E(x))}\\
    f_s &= concat(\mu({\phi_1(y)}),..,\mu(\phi_4(y)),\sigma({\phi_1(y)}),..,\sigma(\phi_4(y)))
\end{align}
where $E$ represents the forward pass of StyleFlow, $\phi_1$ to $\phi_4$ are middle layers ($relu1\_1,relu2\_1,relu3\_1,relu4\_1$) of pre-trained VGG19 encoder.\\[1ex]
\noindent\textbf{Module Details}: 
Unlike AdaIN\cite{huang2017arbitrary} which directly computes $\mu$ and $\sigma$ from the target image to transfer source features, our affine parameters are based on both source and target images. We argue that even for the same reference target image, different source images should use different affine parameters to help transferred images retain special content features of source images, we name them \textit{content-guided affine (CGA) parameters}. As shown in Figure \ref{fig:overview}-A, mean ($CGA_\mu$) and variance ($CGA_\sigma$) are obtained from source features and style features through a learnable network.

To summarize, given source image feature $f_x$, and style feature $f_s$, we have the transferred image feature $\hat{f_x}$:
\begin{align}
    \label{eq:10}
      \hat{f_x} = SAN(f_x,f_s) = \frac{f_x-\mu(f_x)}{\sigma(f_x)} CGA(f_x,f_s)_{\sigma} + CGA(f_x,f_s)_{\mu}
\end{align}
\noindent\textbf{Proof of Fixed-Content}: 
According to Eq \ref{eq:10}, we have the following equation:
\begin{align}
      \frac{\hat{f_x}-CGA(f_x,f_s)_{\mu}}{CGA(f_x,f_s)_{\sigma}} = \frac{f_x-\mu(f_x)}{\sigma(f_x)}
\end{align}
Since content is defined as the normalized term of image features, we have
\begin{align}
      \hat{f}_{c} = \frac{\hat{f_x}-CGA(f_x,f_s)_{\mu}}{CGA(f_x,f_s)_{\sigma}} = \frac{f_x-\mu(f_x)}{\sigma(f_x)} = f_{c}
\end{align}
Therefore, the content of the transferred image feature $\hat{f}_c$ is equal to the content of source image feature $f_c$, which proves that our SAN module performs content-fixed feature transformation.

\subsection{Loss Function}
\label{sec:Loss}
Our loss function can be expressed as:
\begin{align}
      L = L_c + \lambda L_{as} + L_{smooth}
\end{align}
where $L_c$ is the content loss, $L_{as}$ is our proposed aligned-style loss, and 
$\lambda$ is a weighting factor used to trade-off between content and style. $L_{smooth}$ is used to make the generated images smoother.\\[1ex]
\noindent\textbf{Content Loss}:
Following \cite{huang2017arbitrary}, the content loss is defined as the Euclidean distance between the channel-wise normalization of VGG features for the generated image $\hat{x}$ and the source image $x$.
\begin{align}
      L_{c} = \left \|norm(\phi(\hat{x})) - norm(\phi(x))\right \|_2
\end{align}
where $\phi$ refers to the layer $relu\ 4\_1$ of a pre-trained VGG-19 encoder, $norm$ denotes the channel-wise normalization.\\[1ex]
\noindent\textbf{Aligned-Style Loss}:
Considering that the semantic information extracted from VGG-19 of the source image and the target image are not exactly matched in unpaired translation, we extend the style loss in \cite{huang2017arbitrary} to be \textit{aligned-style loss} by setting a parameter $k$ to adjust the percentage of extracted tensors that are used for loss computation. We define $S$ as an ascending sort function. Given a source image $x$, a target image $y$, and the transferred image $\hat{x}$, with an energy function $E(\phi_i(\hat{x}),\phi_i(y))=\left \|\mu(\phi_i(\hat{x}))-\mu(\phi_i(y)) \right \|_2$, where $\phi_i$ ($i\in L=\{1,2,3\}$) represents a set of pre-trained VGG-19 layers $\{relu1\_1,relu2\_1,relu3\_1\}$, we could have the chosen index:
\begin{align}
      C= \{c \in \mathbb{N}_{S}|c\leqslant kN,0<k\leqslant 1 \}
\end{align}
where $\mathbb{N}_{S}$ is the set of indexes of the sorted tensor $S(E(\phi_i(\hat{x}),\phi_i(y)))$, $N$ denotes its total length, and $k$ is the weighting parameter. Therefore,
\begin{align}
      L_{as} = \sum_{i=1}^L\sum_{j\in C}\left \|\mu(\phi_i(\hat{x})_j) - \mu(\phi_i(y)_j)\right \|_2 + \sum_{i=1}^L\sum_{j\in C}\left \|\sigma(\phi_i(\hat{x})_j) - \sigma(\phi_i(y)_j)\right \|_2
\end{align}
where $\phi_i(x)_j$ denotes the $j^{th}$ channel of the output tensor of the $i^{th}$ layer from the set  $\{relu1\_1,relu2\_1,relu3\_1\}$ of a pre-trained VGG-19 encoder.\\[1ex]
\noindent\textbf{Smooth Loss}:
To make the generated image $\hat{x}$ smooth and natural as the target image $y$, we use the Structure-Aware Smoothness Loss proposed in Retinex-Net\cite{wei2018deep} to improve visual coherence during translation.
\begin{align}
      L_{smooth} = ||\nabla\hat{x} \exp(-\lambda_{smooth}\nabla y)||
\end{align}
where $\nabla$ indicates the gradients in both horizontal and vertical axes. $\lambda_{smooth}$ is the weight to adjust the smoothness. During the experiments, $\lambda_{smooth}$ is set to $10$ by default.

\section{Experiments}
To demonstrate the effectiveness of our method in strongly and normally constrained tasks, we use style transfer and photo-realism synthesis to show comparison between our proposed StyleFlow and other state-of-the-art methods of respective fields in this section. More results of various tasks can be found in supplementary materials.
\subsection{Experimental Settings}
As shown in the Figure \ref{fig:overview}, our model consists of $M$ blocks, each block consists of $N$ flows. We use 2 blocks, and 15 flows in each block in the following experiments. The loss weight $\lambda$ is set to 0.1. For style transfer task, the model was trained on single GTX 1080Ti 11G, with an Adam optimizer , an initial learning rate of 5e-5, and a batch size of 1, and $k=\{0.7,0.8,0.9\}$, for 100k iterations. For photo-realism task, the model was trained on two Tesla V100 32G, with an Adam optimizer , an initial learning rate of 5e-5, and a batch size of 1, and $k=\{0.6,0.7,0.8,0.9\}$, for 100k iterations.

\subsection{Style Transfer [Normally Constrained]}
Style transfer aims at transferring the artistic style from a reference image to a content image. This task provides a direct measure of the network's translation capability. We compare our method with the state-of-the-art universal style transfer algorithms, i.e. AdaIN\cite{huang2017arbitrary}, WCT\cite{li2017universal}, StyleSwap\cite{chen2016fast} and ArtFlow\cite{an2021artflow}.\\[1ex]
\noindent\textbf{Dataset}: Following previous art-style transfer works, we use MS-COCO\cite{lin2014microsoft} dataset and Wiki-Art\cite{wikiart} dataset in our experiment. The MS-COCO is used as the source domain, while the Wiki-Art is used as the target domain. For training, the images are resized to 300$\times$400. While for evaluation, all synthesized images are resized to 256$\times$256.\\[1ex]
\begin{figure}[!b]
  \centering
  \includegraphics[width=1.0\linewidth]{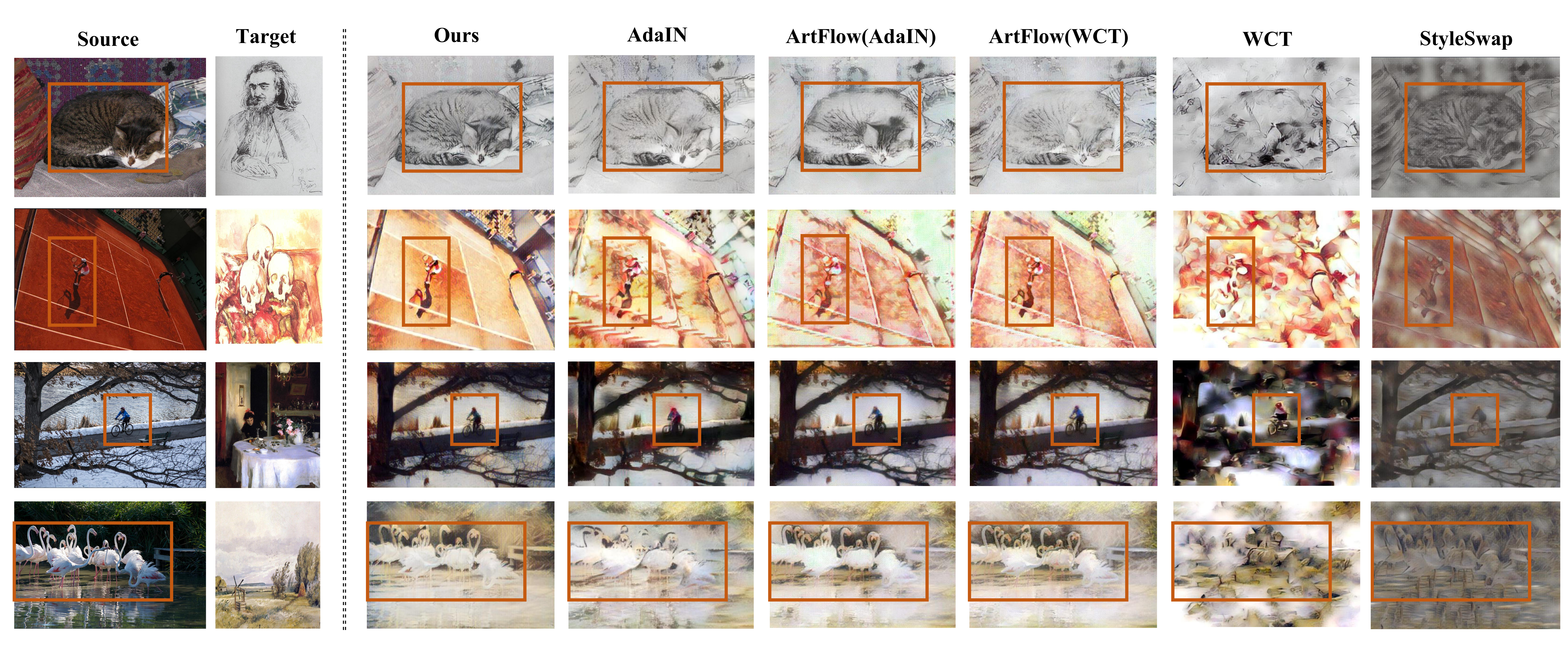}
  \caption{Style transfer results compared with the state-of-the-art style transfer methods.}
  \label{fig:art_pic}
\end{figure}
\noindent\textbf{Qualitative Evaluation}:
We show visual comparisons among different methods in Figure \ref{fig:art_pic}. WCT and StyleSwap can generate stylized images but with severe content distortion. AdaIN preserves content information to a certain extent but content distortion still exist when zoom into details. ArtFlow uses an invertible architecture similar to ours, but it directly perform feature transformation with AdaIN or WCT, which results in certain level of content loss in texture details. Our proposed SAN module performs content-fixed stylization to avoid this problem. As shown in the Figure \ref{fig:art_pic}, our results not only have great stylistic effects, but also achieve the best retention of content information.\\[1ex]
\begin{figure}[!ht]
  \centering
  \includegraphics[width=0.8\linewidth,height=0.46\linewidth]{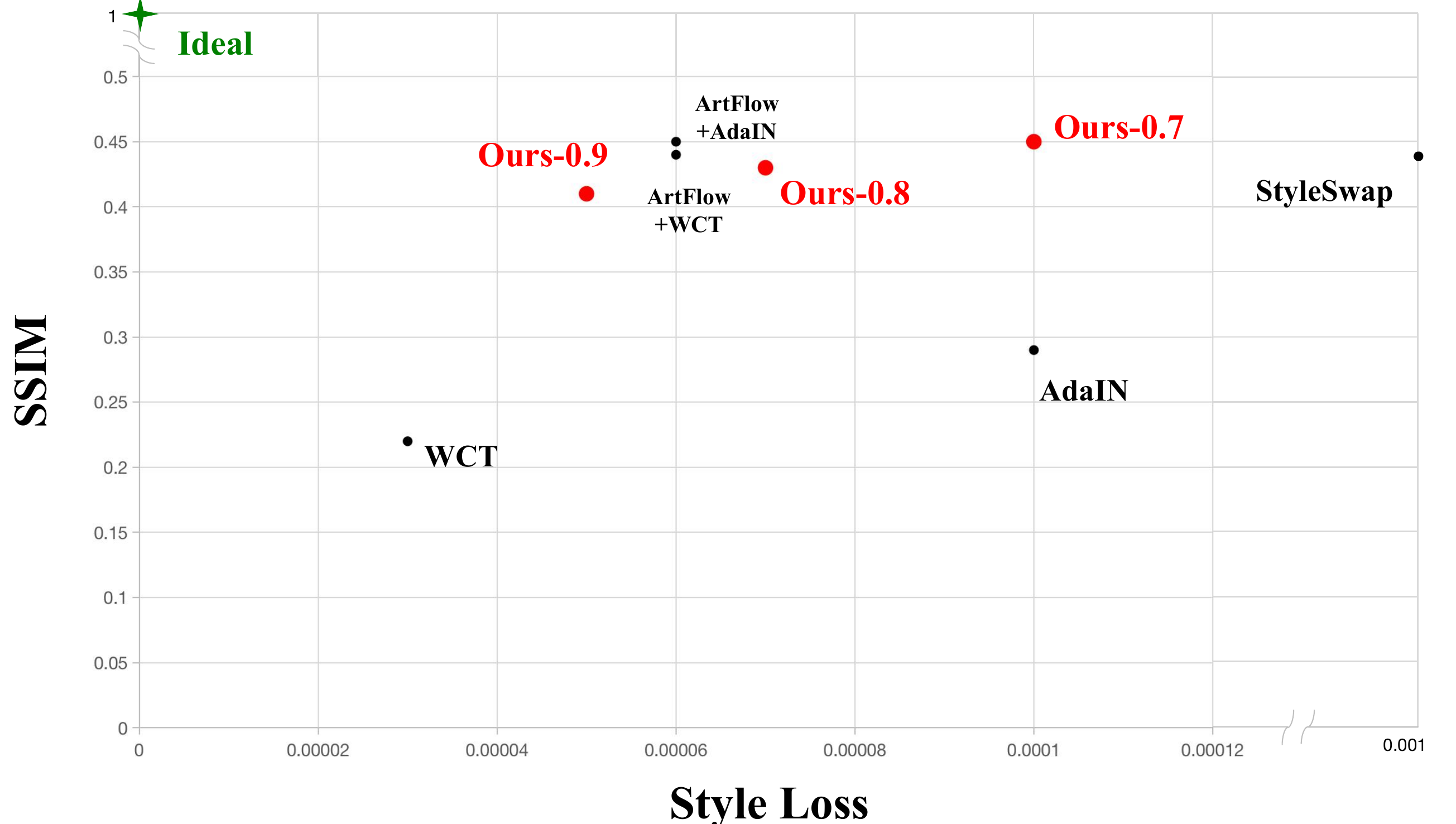}
  \caption{Quantitative results of art-style transfer. SSIM index (higher is better) versus Style Loss (lower is better). Ideal case for art-style transfer is at the top-left corner (\textcolor{green}{green star} †). \textcolor{red}{Red dots} indicate our results.}
  \label{fig:style}
\end{figure}
\noindent{\textbf{Quantitative Evaluation}}:
Following \cite{Hong_2021_ICCV}, we evaluates the stylized images quantitatively using SSIM and style loss, where Structural Similarity Index (SSIM) indicates the performance of content preservation, style loss measures the similarity between the transferred image and the target image. As shown in the Figure \ref{fig:style}, our model with $k=0.9$ achieves the best performance. As the $k$ increases, our model could have better style loss score, but would have a trade-off in content preservation.
\subsection{Photo-Realism [Strongly Constrained]}
\label{sim2real}
Simulation environments have been widely used in many fields. Synthesizing realistic images from simulated environments is a very challenging task. Photo-realism synthesis aims to translate simulated images to realistic images. In this task, we select several state-of-the-art generic I2I models and a specific photo-realism model EPE\cite{richter2021enhancing} for comparison.\\[1ex]
\noindent\textbf{Dataset}: 
We use GTA V dataset\cite{richter2016playing} as the source images, and use Cityscapes\cite{cordts2016cityscapes} and KITTI\cite{geiger2015kitti} for real images. For translation from GTA to Cityscapes, all the images are resized to 800$\times$1000. For translation from GTA to KITTI, images from KITTI are croped with 300$\times$400, and then resized to 800$\times$1000, in order to get better alignment in image resolution.\\[1ex]
\begin{figure*}
  \centering
  \includegraphics[width=1.0\linewidth]{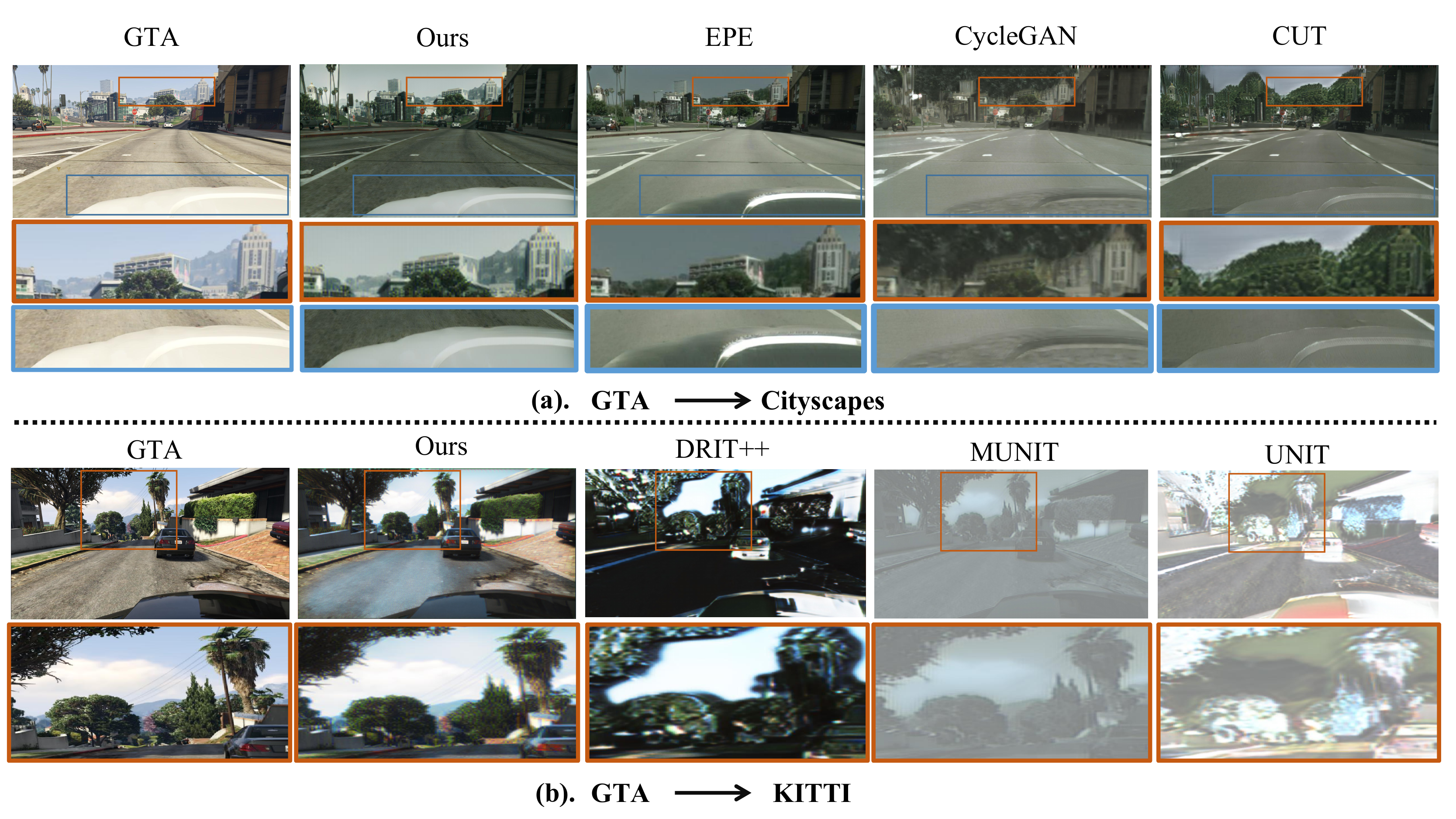}
  \caption{Our results compared with several baselines. As shown in the \textcolor{orange}{orange} and \textcolor{blue}{blue} box, our method greatly preserves the content information. (a) Translation from GTA to Cityscapes. EPE\cite{richter2021enhancing}, CycleGAN\cite{engin2018cycle}, and CUT\cite{park2020contrastive} turn the color of the vehicle from white to black and generate trees in the sky. (b) Translation from GTA to KITTI. Both DRIT++\cite{lee2020drit++} and UNIT\cite{liu2017unsupervised} fail to keep the semantic information. MUNIT\cite{zhu2017multimodal} makes everything blurry.}
  \label{fig:gta}
\end{figure*}
\noindent\textbf{Qualitative Evaluation}: As shown in the Figure \ref{fig:gta}, all previous I2I models fail to retain full content information, different levels of content distortion exist. EPE uses an auxiliary segmentation task to help keep the content. However, it still hallucinates trees on building/sky. Our model achieves content-fixed translation without extra information. (More details are shown in the orange box and blue box in the Figure \ref{fig:gta}).\\[1ex]

\begin{table}[!t]
  \centering
   \caption{Comparison with the previous work on translation from GTA to Cityscapes and GTA to KITTI. SSIM(higher is better), KID(lower is better), and FID(lower is better) are used for evaluation. Data in \textbf{bold} indicates the best performance. }
  \label{tab:example1}
\resizebox{\textwidth}{7mm}{
\begin{tabular}{|l|l|l|l||l|l|l|l||l|l|l|l|}  
  \hline  
  Dataset & Method & Ours-0.9 & Ours-0.8 & Ours-0.7 & Ours-0.6 & CycleGAN\cite{engin2018cycle} & MUNIT\cite{zhu2017multimodal} & DRIT++\cite{lee2020drit++} & CUT\cite{park2020contrastive} & UNIT\cite{liu2017unsupervised} & EPE\cite{richter2021enhancing} \\ \hline  
  GTA$\rightarrow$ Cityscapes & SSIM$\uparrow$ & 0.78 & 0.81 & 0.84 & \textbf{0.87} & 0.71 & 0.60 & 0.14 & 0.71 & 0.22 & 0.8 \\  
   & KID$\downarrow$ & 0.0432/0.0048 & 0.0451/0.0053 & 0.0467/0.0057 & 0.0566/0.0058 & 0.0483/0.0044 & 0.0868/0.0067 & 0.1113/0.0061 & \textbf{0.0390/0.0044} & 0.0908/0.0073 & 0.0409/0.0046 \\  
   & FID$\downarrow$ & \textbf{0.44} & 0.47 & 0.48 & 0.58 & 12.51 & 18.7 & 3.03 & 5.04 & 11.34 & 0.89\\ \hline  
   GTA$\rightarrow$ KITTI & SSIM$\uparrow$ & 0.72 & 0.74 & 0.78 & \textbf{0.82} & 0.76 & 0.58 & 0.08 & 0.63 & 0.11 & /\\  
   & KID$\downarrow$ & \textbf{0.0714/0.0056} & 0.0748/0.0060 & 0.0775/0.0077 & 0.0809/0.0070 & 0.0837/0.0087 & 0.1015/0.0084 & 0.1382/0.0102 & 0.0832/0.0097 & 0.1352/0.0103 & /\\  
   & FID$\downarrow$ & \textbf{3.30} & 3.48 & 3.86 & 3.95 & 11.15 & 37.52 & 4.94 & 5.16 & 10.82 & /\\
  \hline  
  \end{tabular}}
\end{table}
\noindent\textbf{Quantitative Evaluation}:
We quantitatively compare StyleFlow(with $k=\{0.6,0.7,0.8,0.9\}$) with five baselines in two datasets. Following previous methods, we used SSIM (Structural Similarity Index), FID (Frechet Inception Distance) and KID (Kernel Inception Distance) as evaluation metrics. Same as the previous experiment, the SSIM is used to measure the performance of content preservation. While FID and KID measure difference between two image distributions. Results in table \ref{tab:example1} indicate the superiority of our model over previous methods.

\section{Discussion}
\begin{figure}
  \centering
  \includegraphics[width=0.5\linewidth,height=0.25\linewidth]{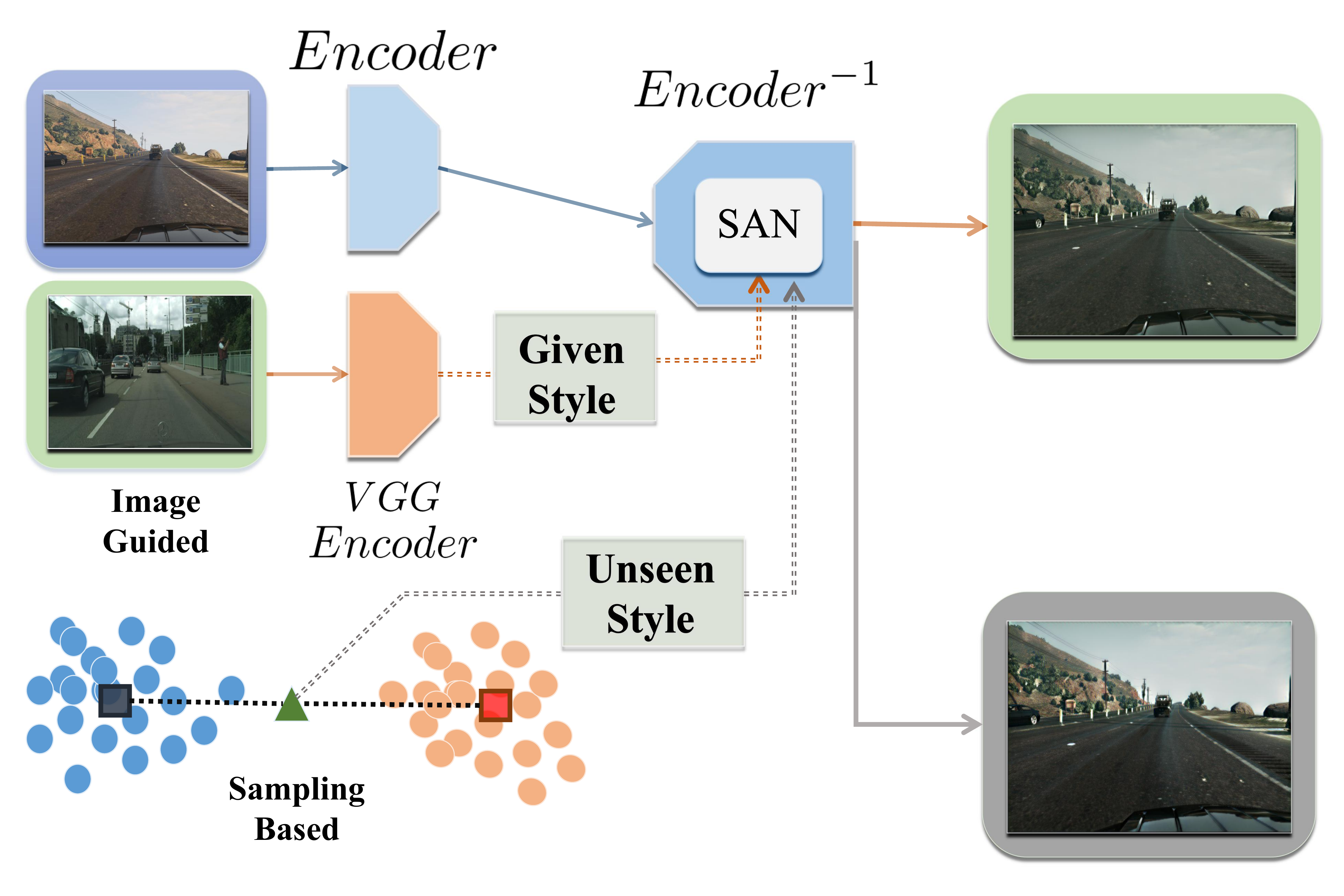}
  \caption{Inference Overview.}
  \label{fig:inf}
\end{figure}
\begin{figure}
  \centering
  \includegraphics[width=0.8\linewidth, height=0.4\linewidth]{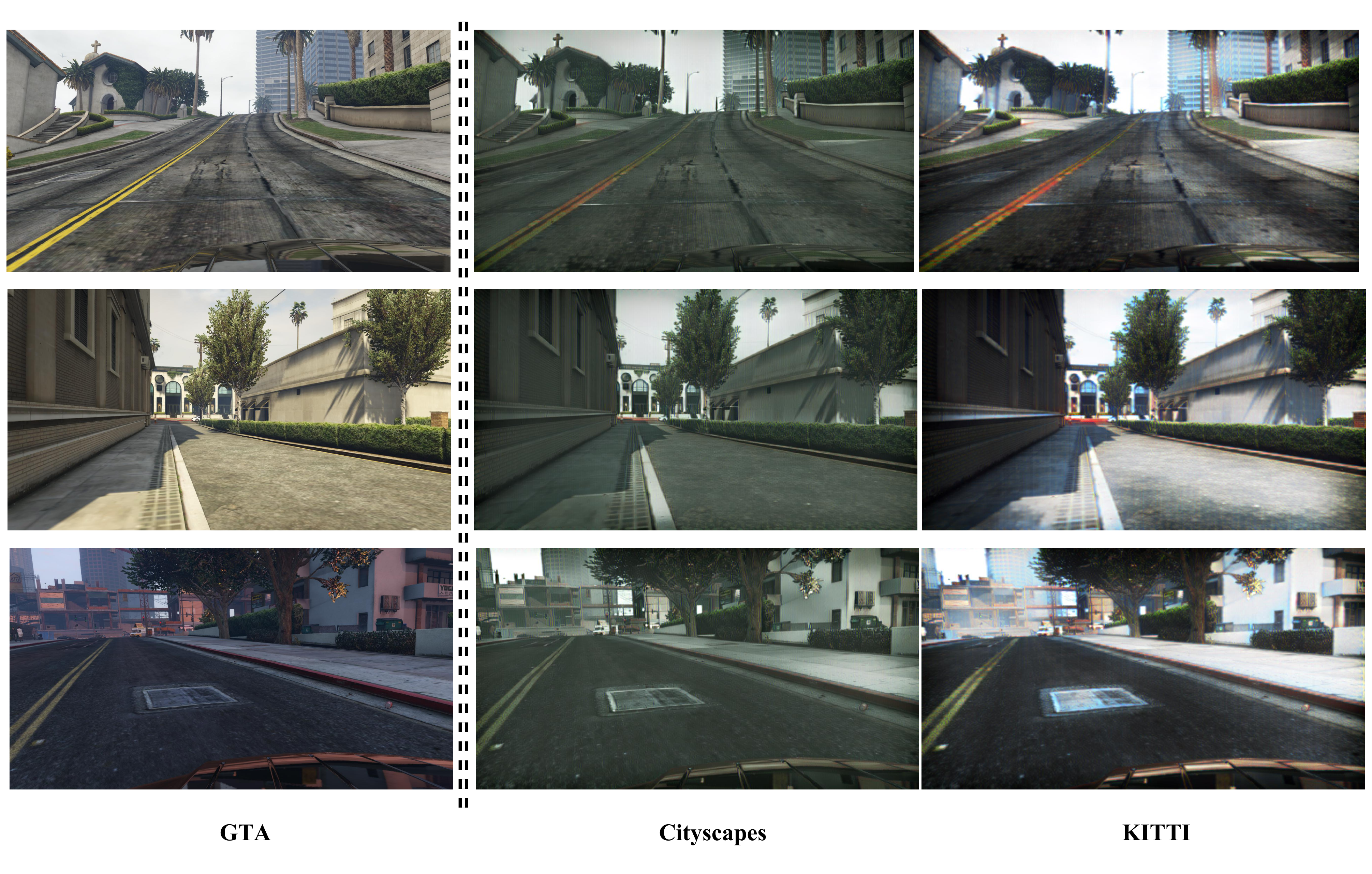}
  \caption{Unpaired multi-domain image-to-image translation from GTA\cite{richter2016playing} to Cityscapes\cite{cordts2016cityscapes} and GTA to KITTI\cite{geiger2015kitti}.}
  \label{fig:multi-domain}
\end{figure}
\begin{figure}
  \centering
  \includegraphics[width=0.9\linewidth,height=0.2\linewidth]{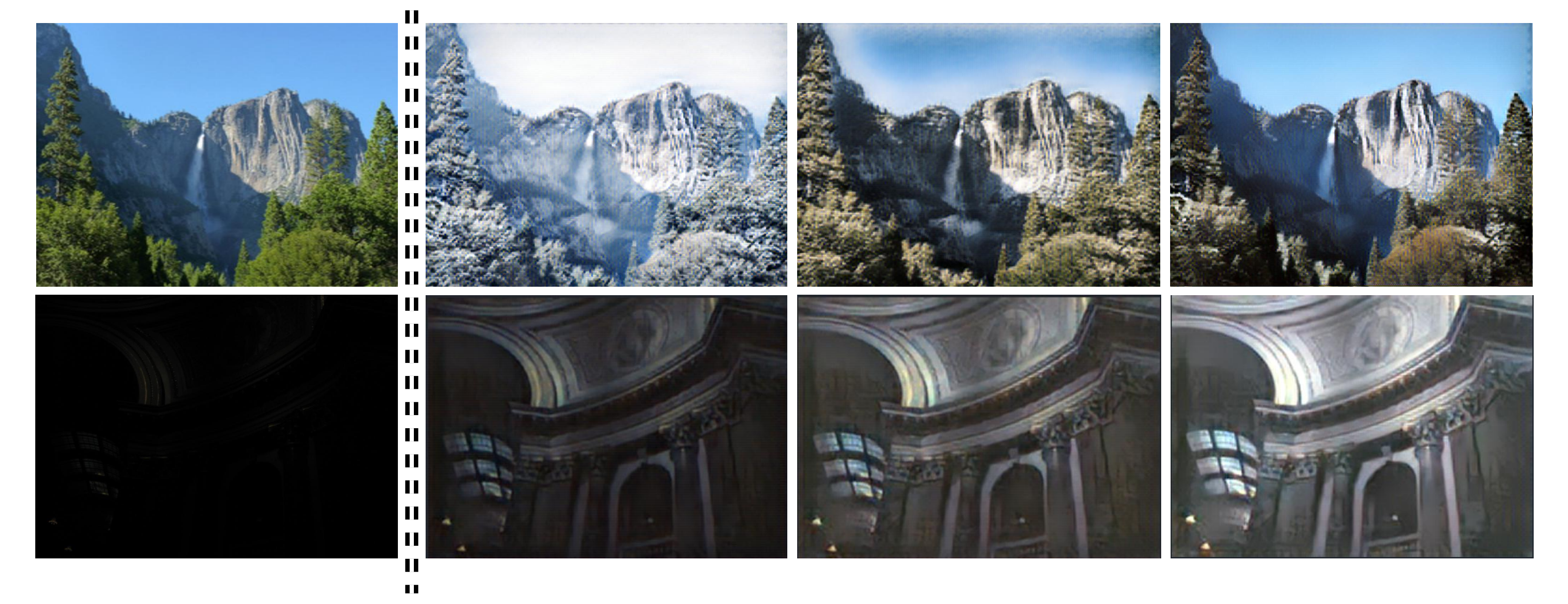}
  \caption{Unpaired diverse image-to-image translation. (\textit{Top}) Multi-modality image to image translation from summer to winter\cite{zhu2017unpaired}. (\textit{Bottom}) Multi-modality image to image translation from dark to light\cite{wei2018deep}}
  \label{fig:multi-modal}
\end{figure}
\begin{figure}
  \centering
  \includegraphics[width=0.915\linewidth,height=0.125\linewidth]{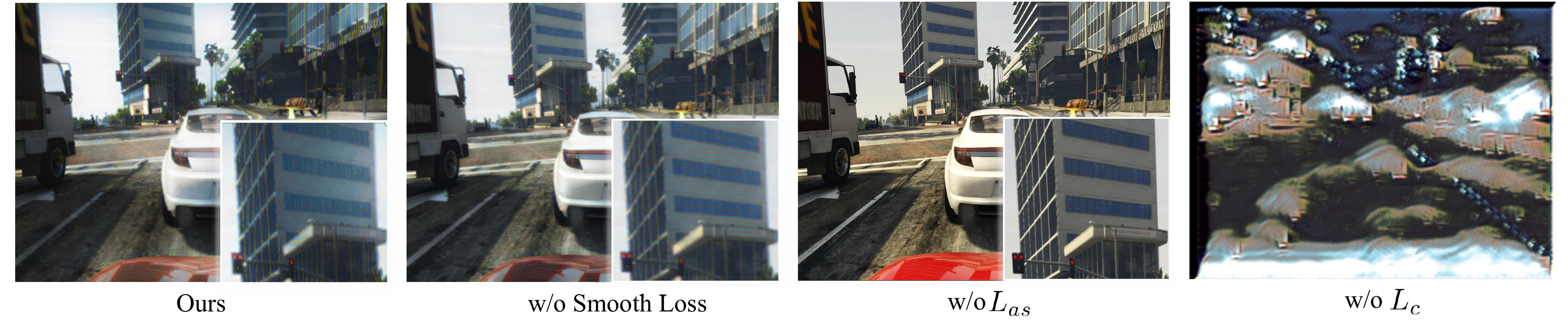}
  \caption{Ablation study of different loss functions.}
  \label{fig:ablation}
\end{figure}
\noindent \textbf{Inference.}
As shown in the Figure \ref{fig:inf}, for inference, our model supports both image-guided image to image translation and sampling-based image to image translation. For sampling-based inference, the style code is sampled from learned latent distribution. Further details are provided in supplementary materials, due to the limitation of paper length.\\[0.5ex]
\noindent \textbf{Multi-domain Image Translation.} In section 
\ref{sim2real}, our model is trained in a multi-domain image translation manner to perform translation from GTA to Cityscapes and GTA to KITTI at the same time. As shown in  Figure \ref{fig:multi-domain}, our model did learn how to transfer source images to different styles for different domains, while maintaining high content preservation for each output.\\[0.5ex]
\noindent \textbf{Multi-modal Image Synthesis.} Figure \ref{fig:multi-modal} indicates the ability for multi-modal synthesis of our method. Training only on a single model, our method effectively generates diverse outputs with the same style properties, and keeps photorealistic generation quality in each output. Top line indicates diverse synthesis from summer to winter. Bottom line indicates translation from dark to different level of brightness. \\[0.5ex]
\noindent \textbf{Ablation Study.}
As shown in the Figure \ref{fig:ablation}, we conduct ablation study to verify the effectiveness of our loss functions, using translation from GTA to KITTI as an example. Without smooth loss, the generated images would have gird-like noises. Without the aligned-style loss ($L_{as}$), the overall visual effect is reduced, which makes the generated images look unrealistic. Content loss ($L_c$) is used to preserve the semantic information, without this term, the generated images can become distorted and over-stylized.\\[0.5ex]
\noindent \textbf{Limitation.}
Although our model outperforms previous methods  in strongly and normally constrained tasks, we failed to achieve admirable results in all weakly constrained translation tasks. Future work includes extending this model to weakly constrained setting.\\[0.5ex]
\noindent \textbf{Broader Impact.}
Our proposed generative model could be used as a tool to further eliminate the gap between simulation and reality, which can be widely used in self-driving and medical areas. The use of image synthesis would not lead to privacy issues, but might create fake news. More efforts are needed in the future to develop regulations to restrict the usage of synthesized data.
\section{Conclusion}
In this paper, we categorize image-to-image translation problem into three tasks: strongly constrained translation, normally constrained translation and weakly constrained translation. We proposed an invertible network StyleFlow, with Style-Aware Normalization (SAN) module for content-fixed image-to-image translation. We have proved that our method can achieve full content preservation during translation. Qualitative and quantitative results show that our model is more suitable for strongly and normally constrained translation, while the previous work have better performance on weakly constrained translation.

\begin{figure*}
  \centering
  \includegraphics[width=1.0\linewidth,height=1.2\linewidth]{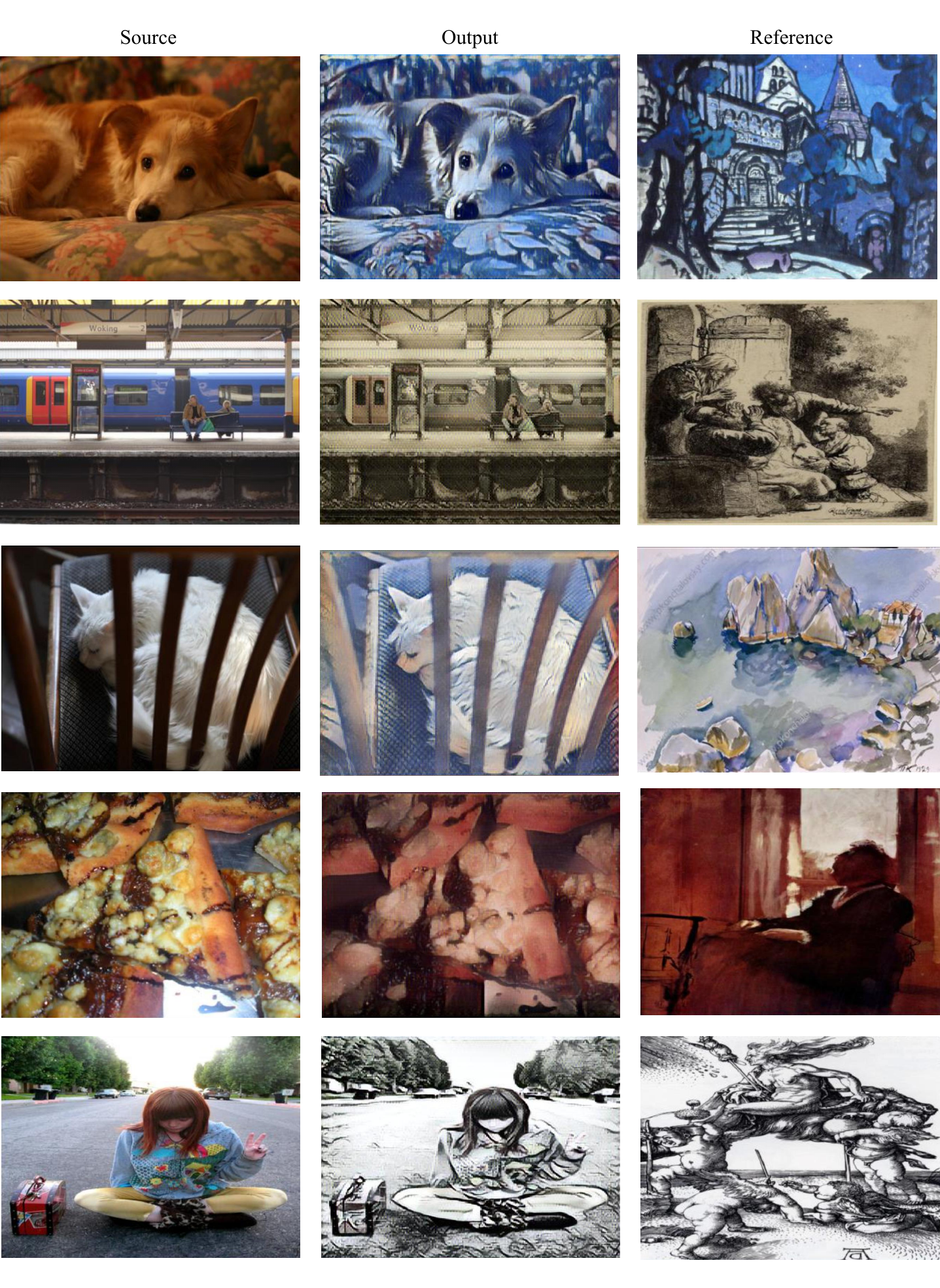}
  \caption{Image-guided style transfer results from MS-COCO\cite{lin2014microsoft} to WikiArt\cite{wikiart}}
  \label{fig:ars}
\end{figure*}

\begin{figure*}
  \centering
  \includegraphics[width=1.0\linewidth,height=1.2\linewidth]{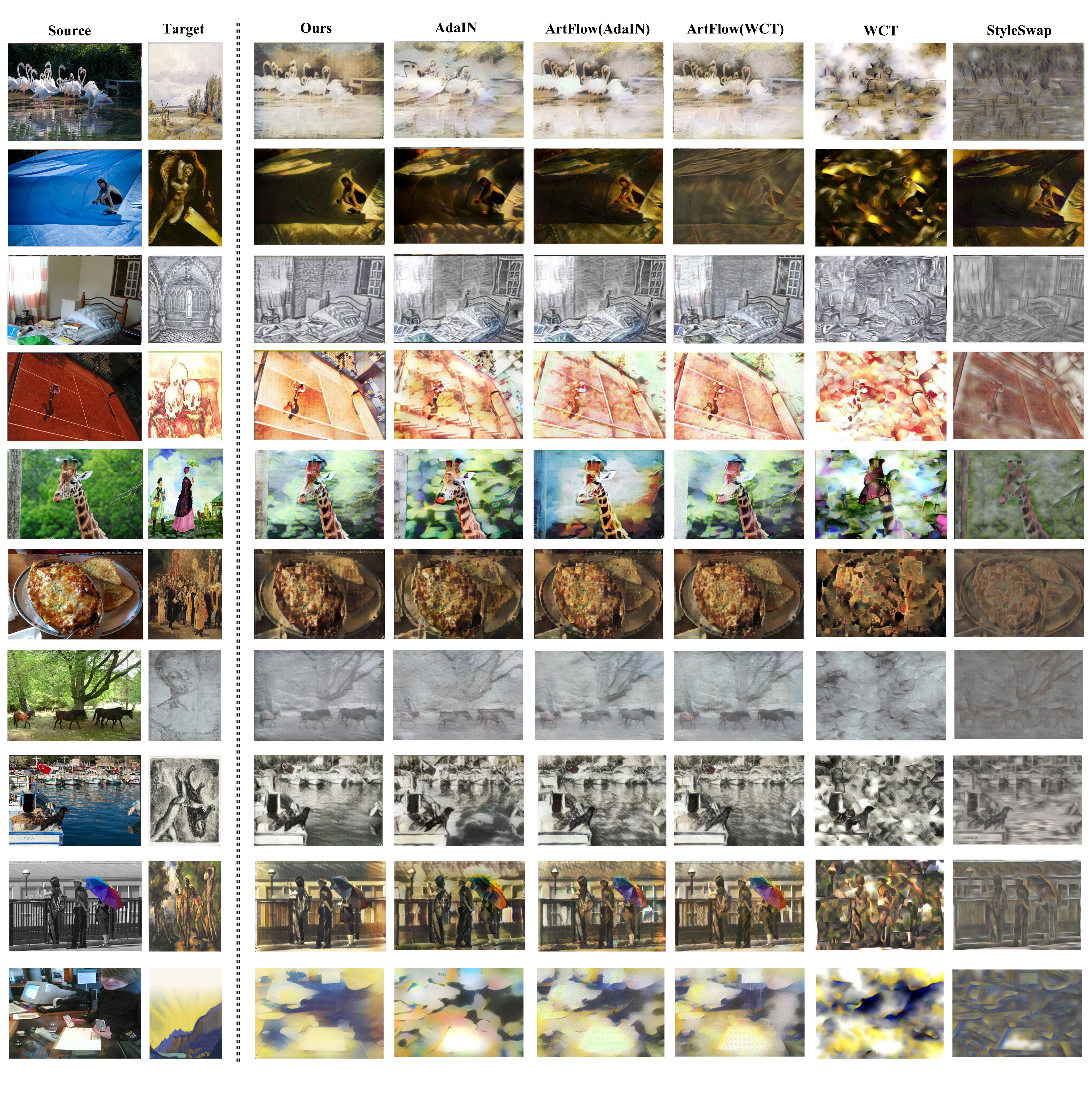}
  \caption{Comparison on style transfer from MS-COCO\cite{lin2014microsoft} to WikiArt\cite{wikiart} between StyleFlow and the previous methods.}
  \label{fig:art_all}
\end{figure*}

\begin{figure*}
  \centering
  \includegraphics[width=1.0\linewidth,height=1.2\linewidth]{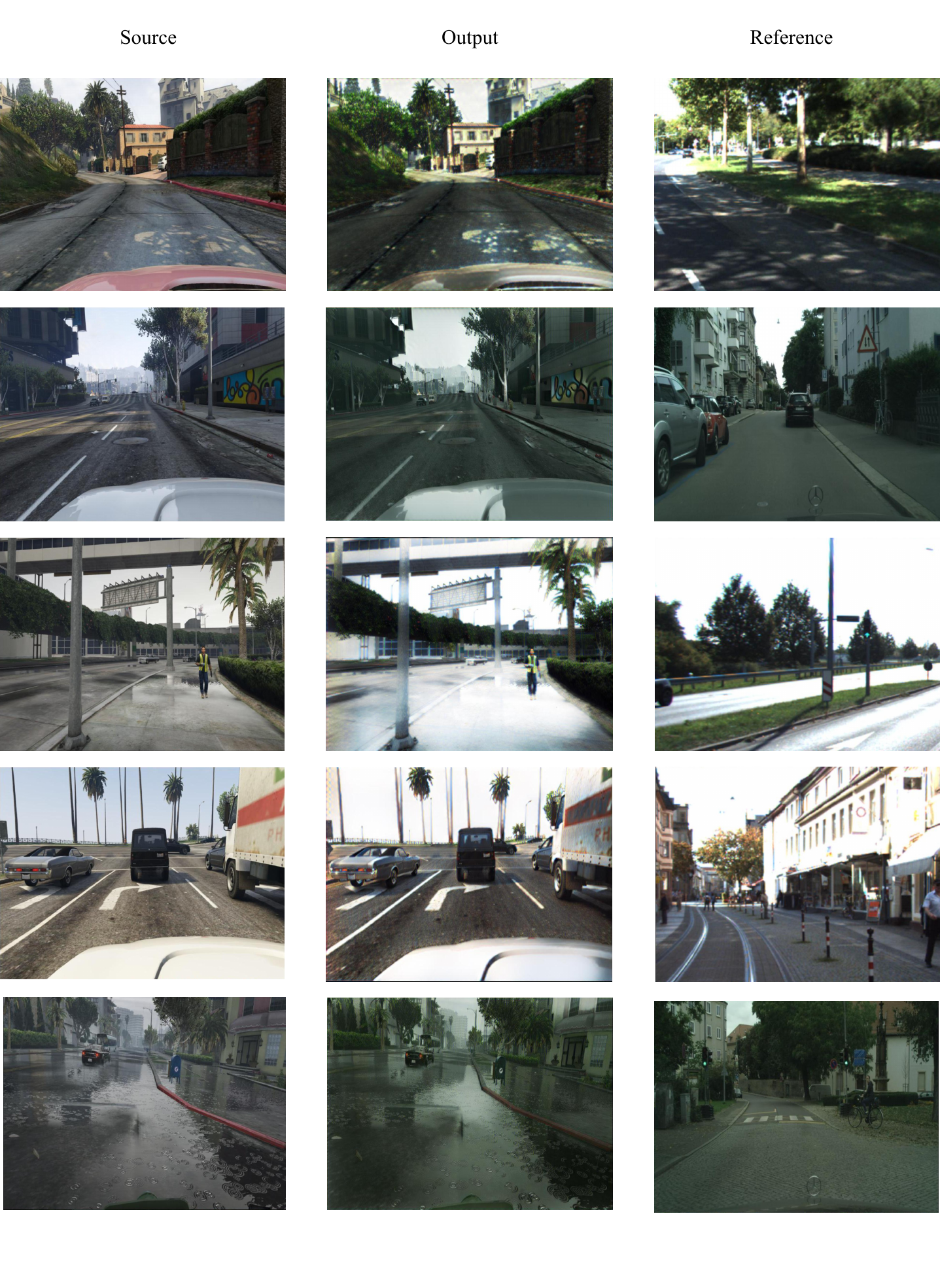}
  \caption{Image guided photo-realism on GTA\cite{richter2016playing} to Cityscapes\cite{cordts2016cityscapes} and GTA to KITTI\cite{geiger2015kitti}.}
  \label{fig:gg}
\end{figure*}

\begin{figure*}
  \centering
  \includegraphics[width=1.0\linewidth,height=1.2\linewidth]{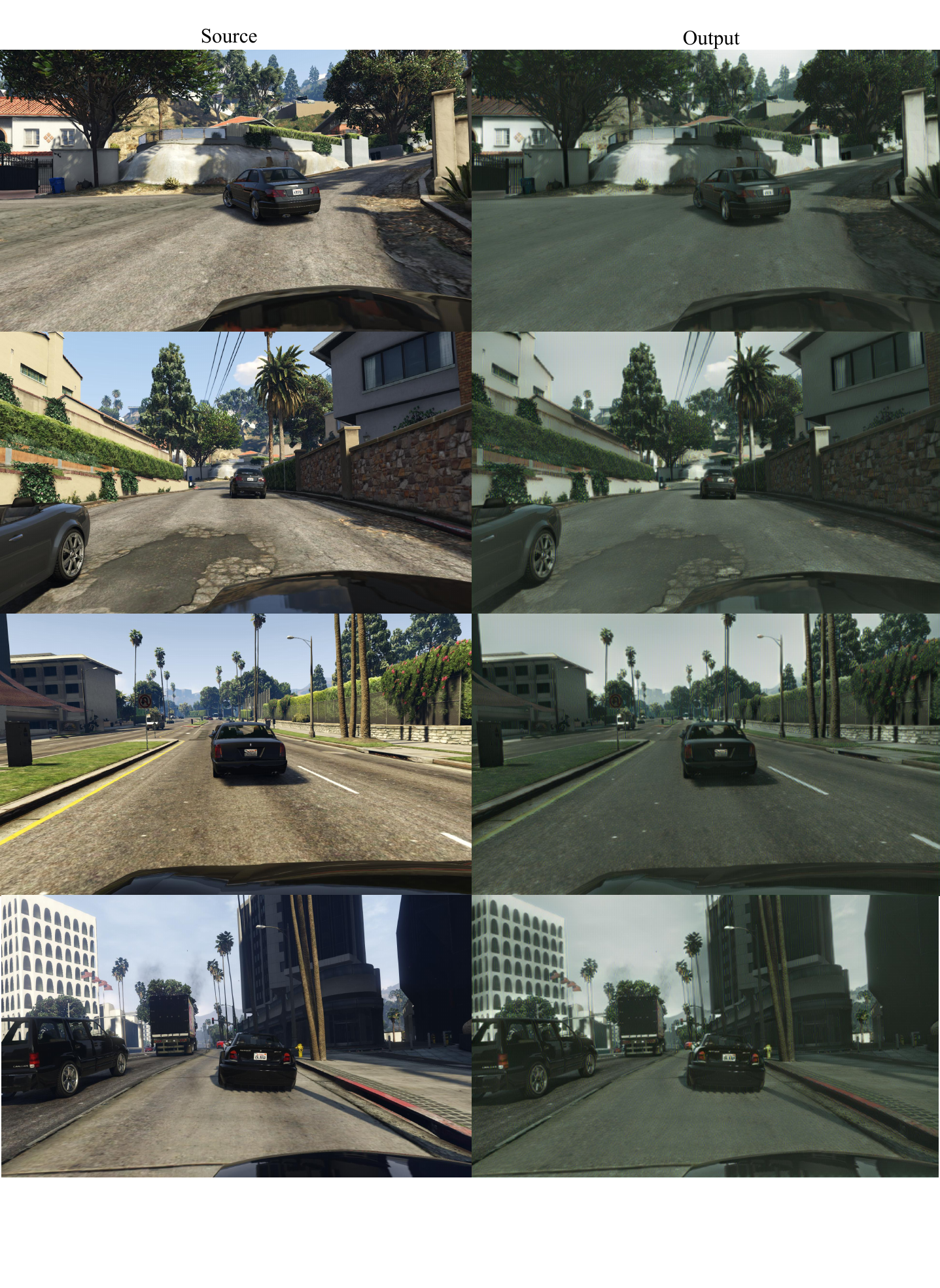}
  \caption{Results on photo-realism from GTA\cite{richter2016playing} to Cityscapes\cite{cordts2016cityscapes} by sampling-based inference.}
  \label{fig:gs}
\end{figure*}

\begin{figure*}
  \centering
  \includegraphics[width=1.0\linewidth]{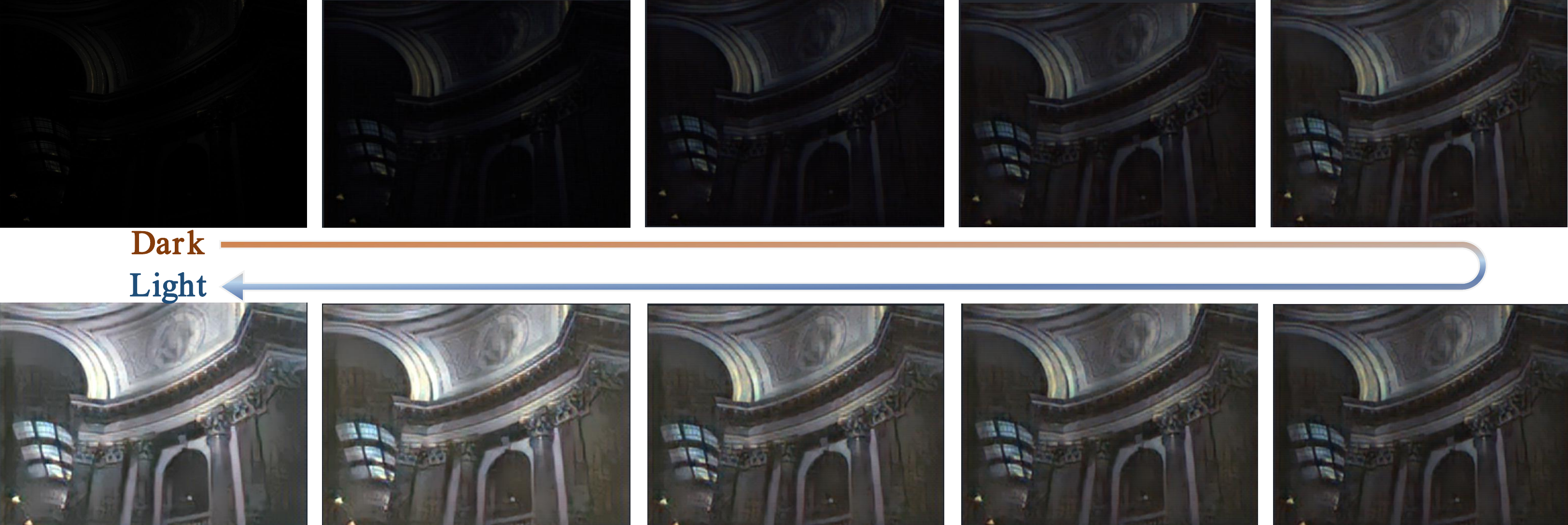}
  \caption{Continuous outputs from one single model for dark-to-light translation.}
  \label{fig:interpolation}
\end{figure*}

\begin{figure*}
  \centering
  \includegraphics[width=1.0\linewidth]{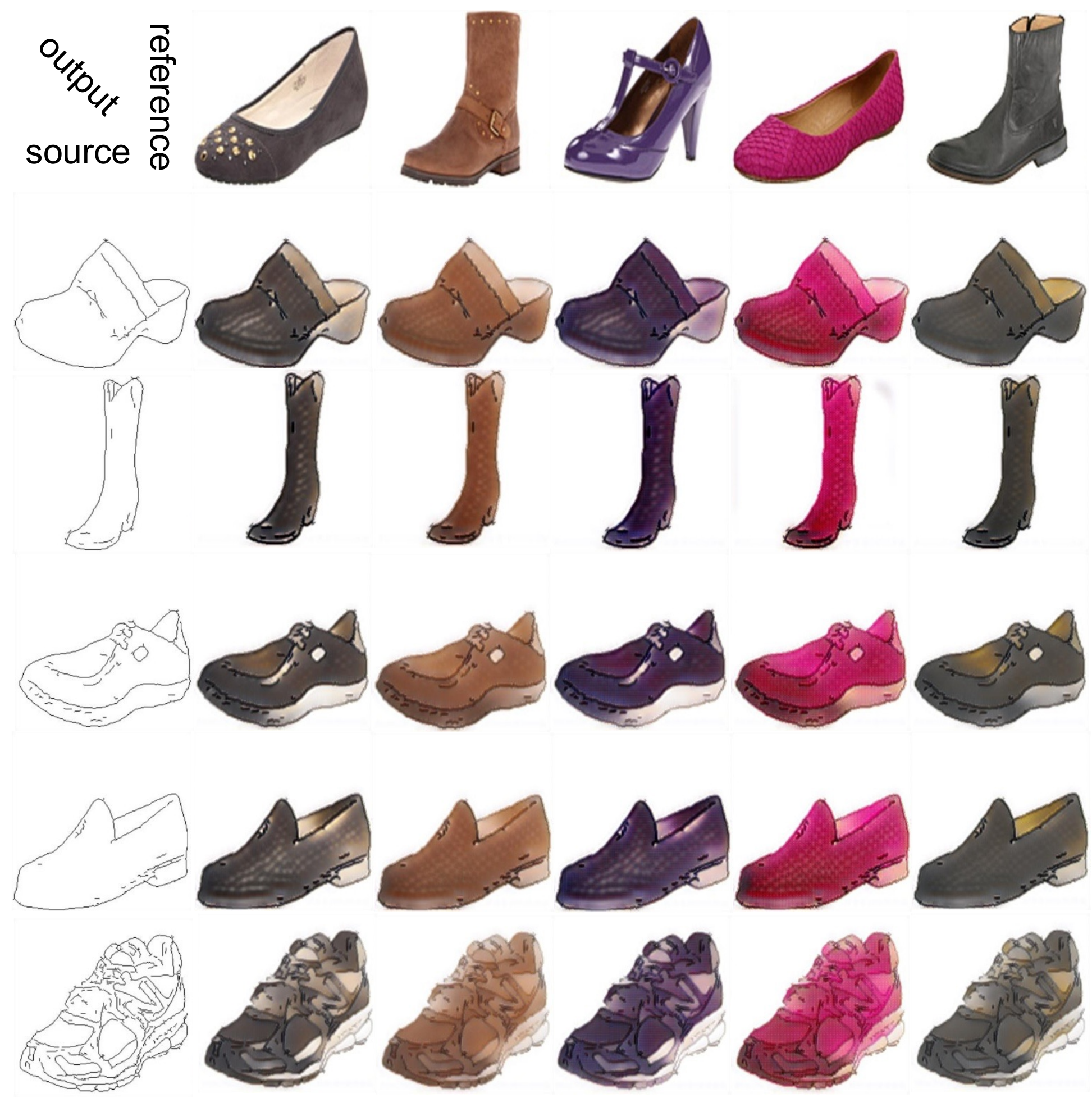}
  \caption{Diverse outputs for edges-to-shoes translation.}
  \label{fig:edge2shoe_multimodal}
\end{figure*}

\begin{figure*}
  \centering
  \includegraphics[width=0.9\linewidth]{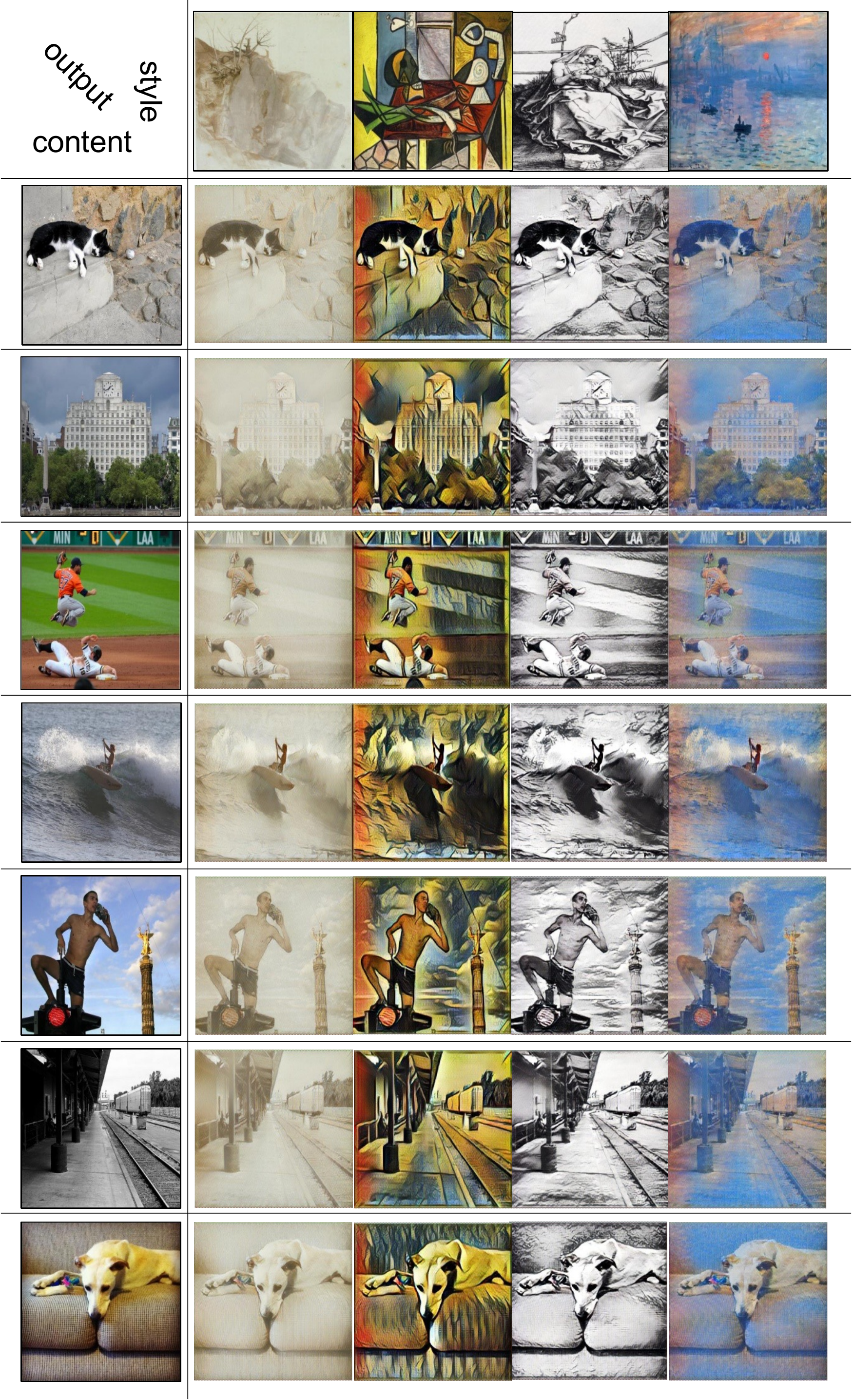}
  \caption{Diverse outputs for arbitrary style transfer.}
  \label{fig:wikiart_multimodal}
\end{figure*}

\begin{figure*}
  \centering
  \includegraphics[width=0.8\linewidth]{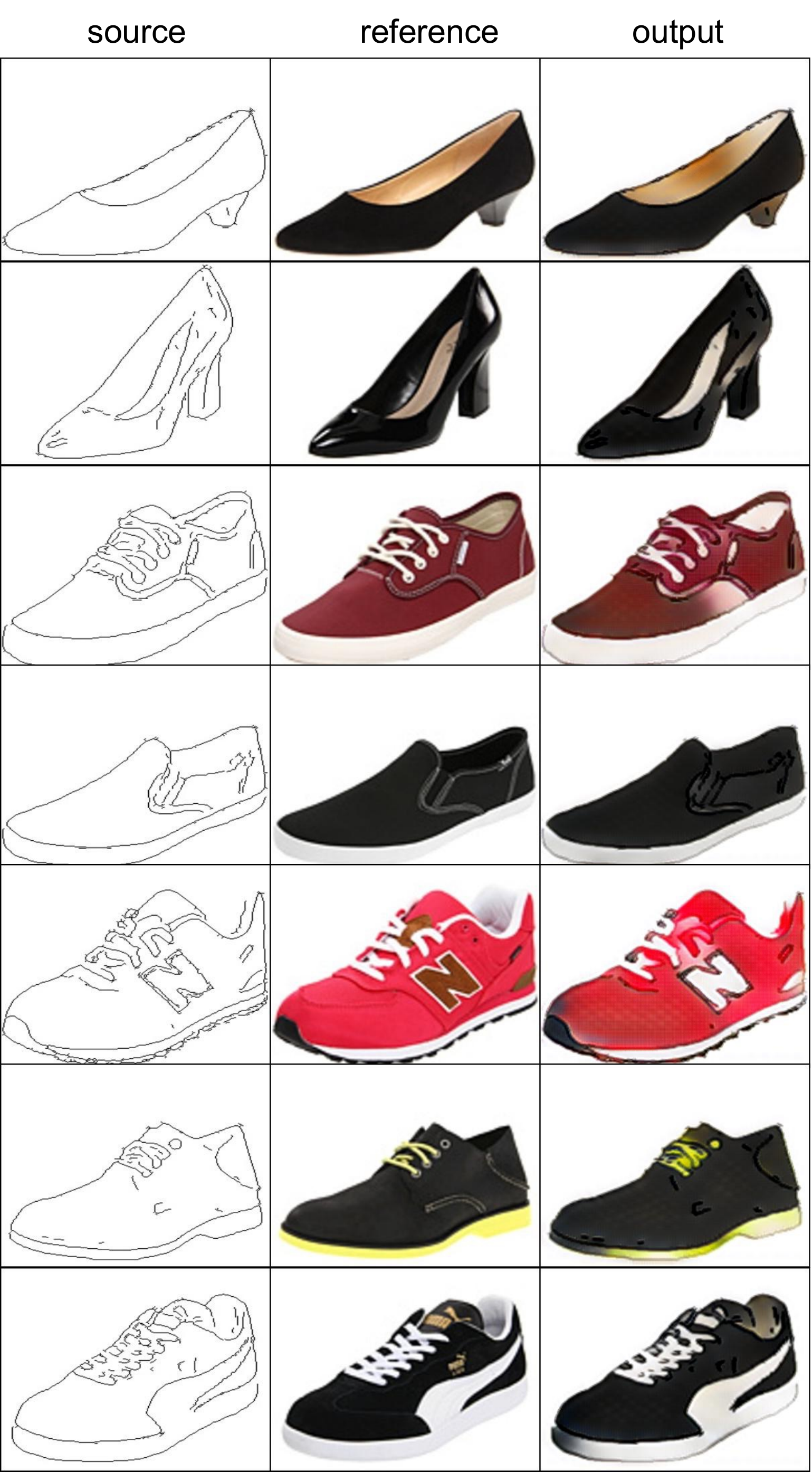}
  \caption{Translation from edges to shoes\cite{xie15hed},\cite{fine-grained}}
  \label{fig:edge2shoe}
\end{figure*}

\begin{figure*}
  \centering
  \includegraphics[width=1.0\linewidth,height=1.2\linewidth]{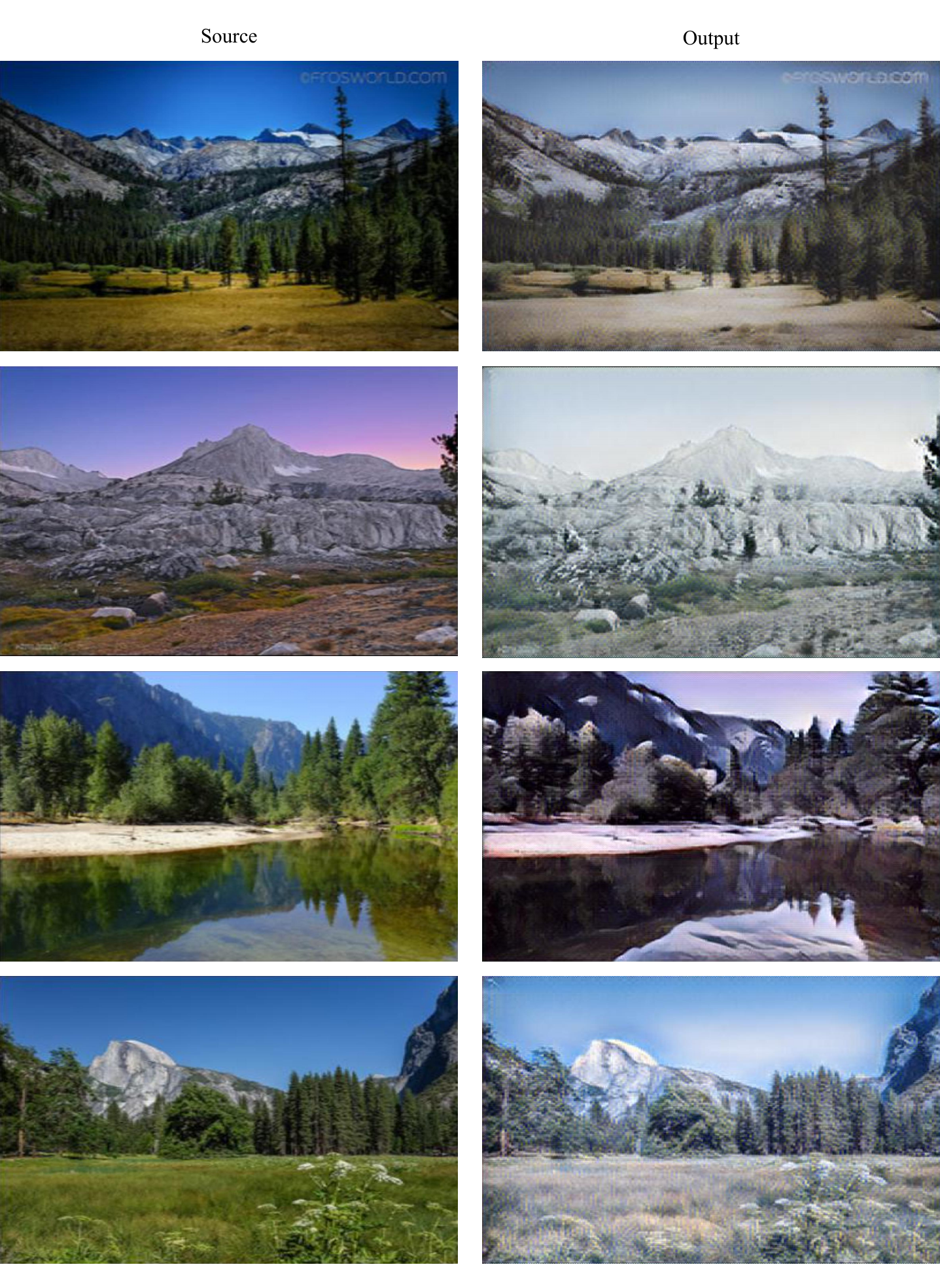}
  \caption{Translation from summer to winter\cite{zhu2017unpaired}.}
  \label{fig:summer}
\end{figure*}

\begin{figure*}
  \centering
  \includegraphics[width=1.0\linewidth,height=1.2\linewidth]{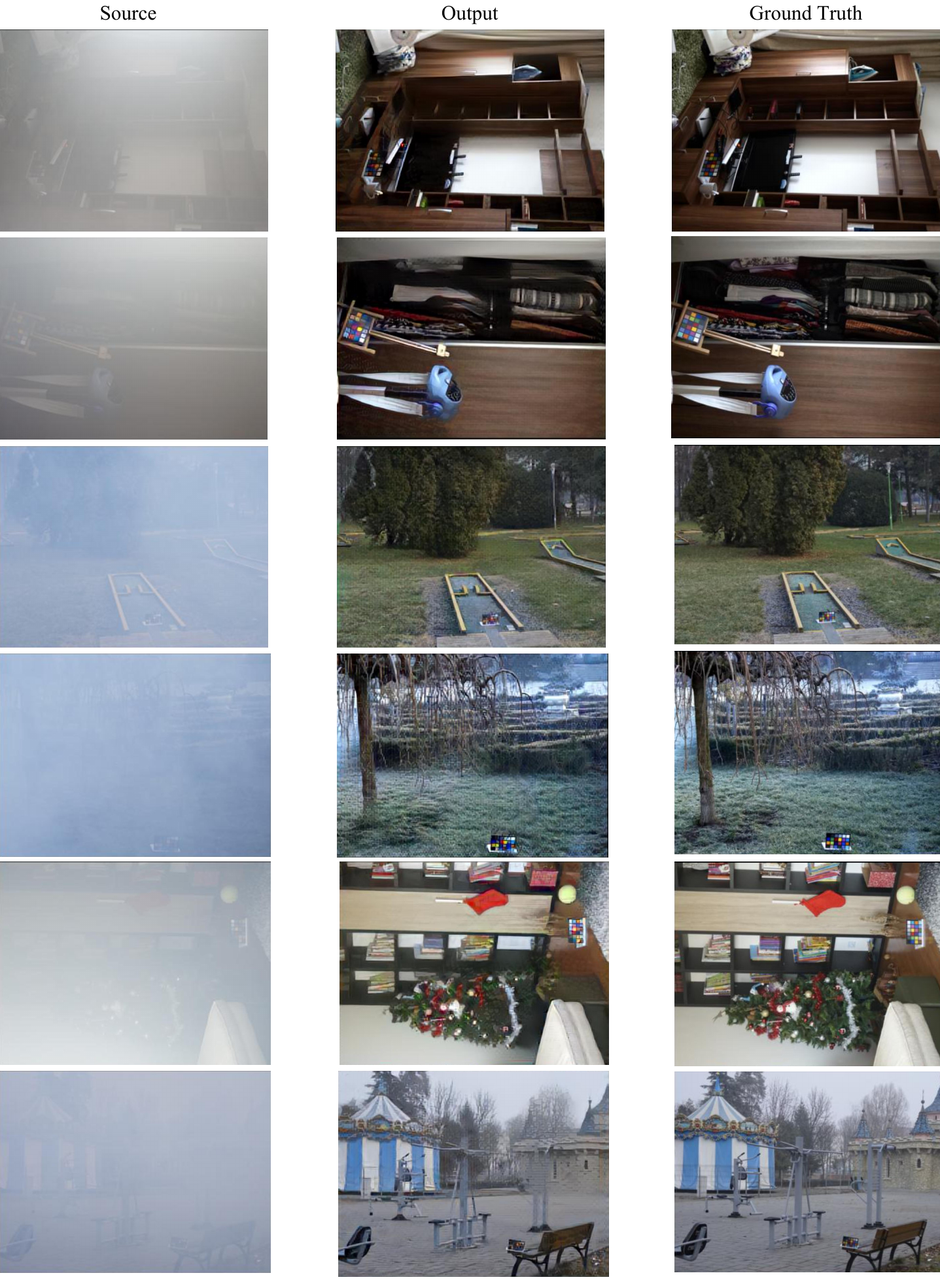}
  \caption{Results on image dehazing.\cite{Dense-Haze_2019}}
  \label{fig:haze}
\end{figure*}

\begin{figure*}
  \centering
  \includegraphics[width=1.0\linewidth,height=1.2\linewidth]{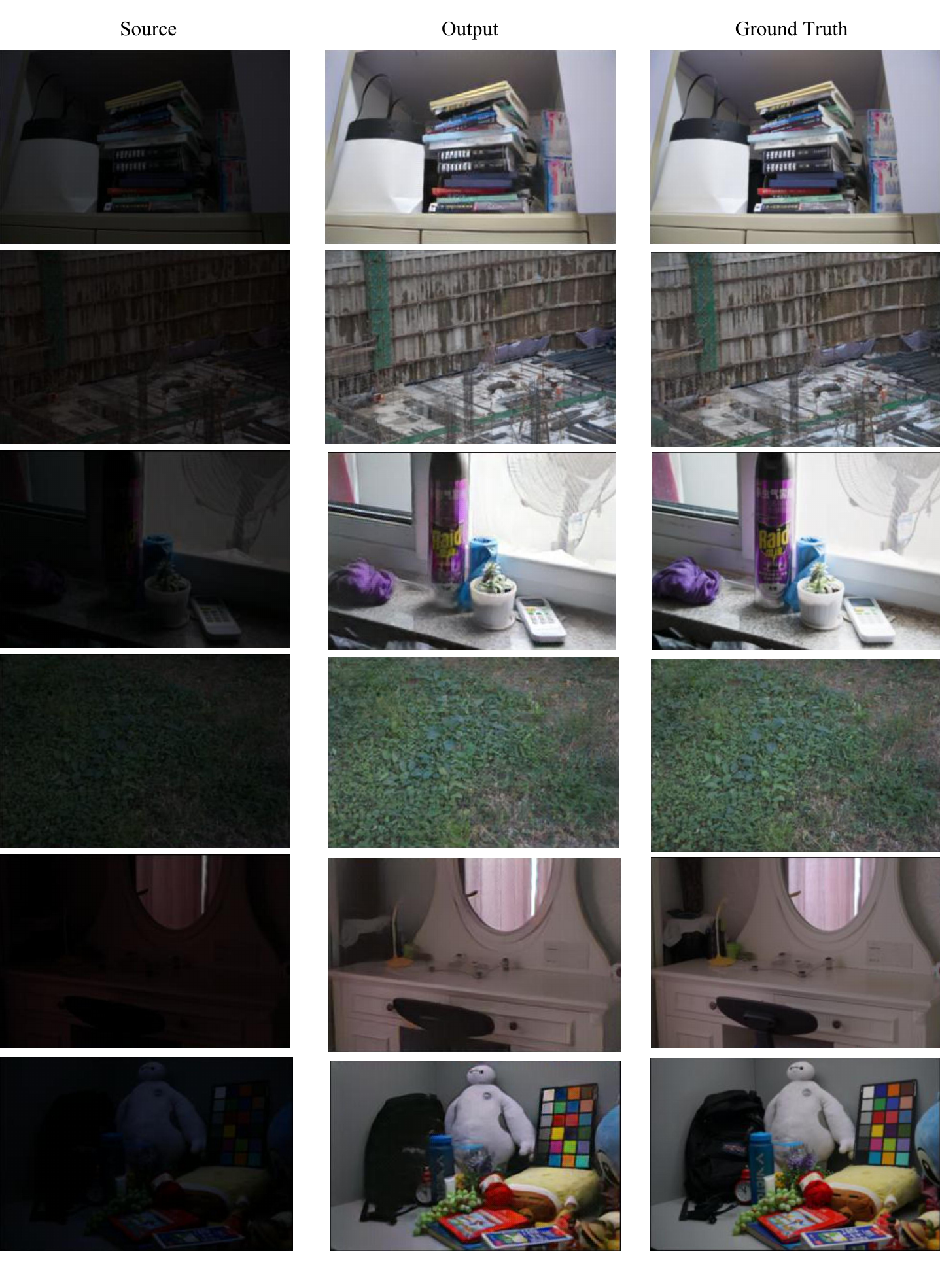}
  \caption{Results on image enhancement\cite{wei2018deep}.}
  \label{fig:light}
\end{figure*}

\begin{figure*}
  \centering
  \includegraphics[width=1.0\linewidth,height=1.2\linewidth]{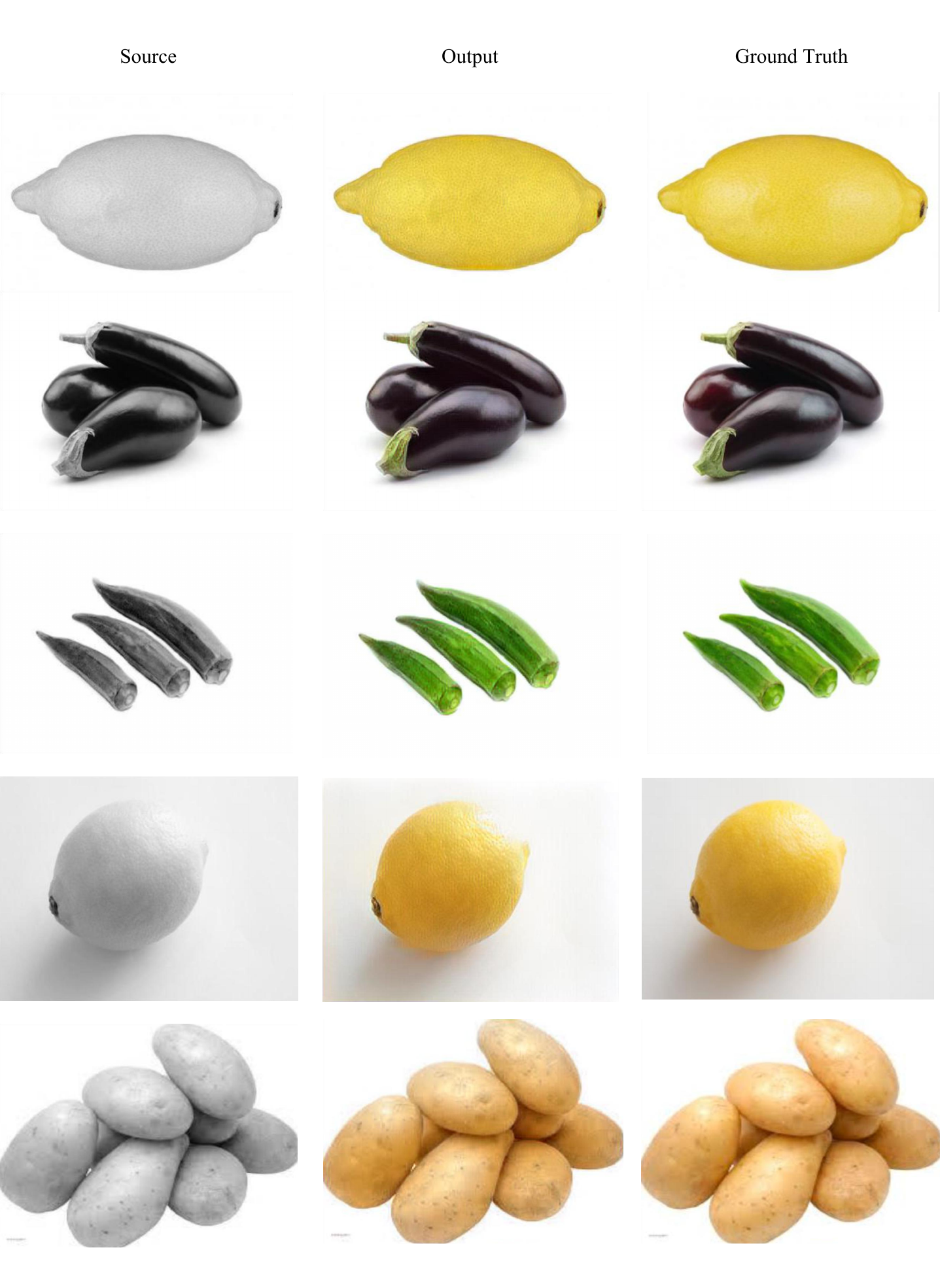}
  \caption{Results on colorization\cite{anwar2020image}.}
  \label{fig:color}
\end{figure*}

\begin{figure*}
  \centering
  \includegraphics[width=1.0\linewidth,height=1.2\linewidth]{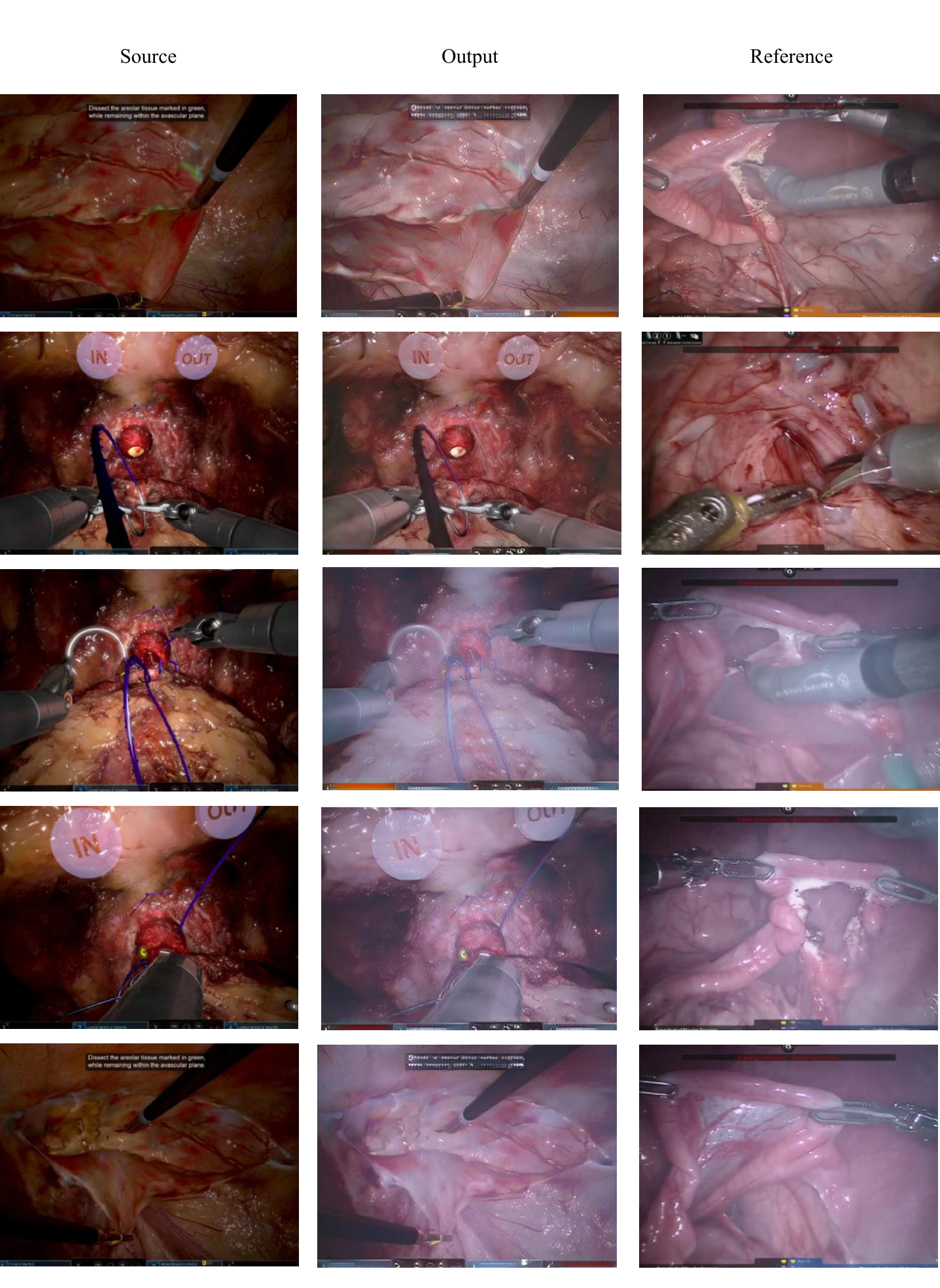}
  \caption{Results on translation from sim to real\cite{zia2021surgical}.}
  \label{fig:sim}
\end{figure*}

\clearpage

%
%
\bibliographystyle{plain}
\bibliography{paper_arxiv}
\end{document}